\documentclass[11pt]{article}

\usepackage[preprint]{acl}

\usepackage{times}
\usepackage{latexsym}
\usepackage{amsmath}
\usepackage{amssymb}
\usepackage{booktabs}
\usepackage{bbm}
\usepackage[T1]{fontenc}
\usepackage{multirow}
\usepackage{nicefrac}

\usepackage[utf8]{inputenc}

\usepackage{microtype}

\usepackage{inconsolata}
\usepackage{cleveref}

\usepackage{graphicx}
\newcommand{\llama}{Llama 3 8B}
\newcommand{\olmo}{OLMo 2 7B}

\title{Rethinking Data Selection: The Importance of Coverage over Difficulty in LLM Fine-Tuning}

\author{Lalchand Pandia \\
Department of Computer Science\\
University of Chicago\\
\texttt{lcpandia@cs.uchicago.edu} \\
\And
Kanishka Misra \\
Department of Linguistics \\
The University of Texas at Austin \\
\texttt{kmisra@utexas.edu} \\
\AND
Allyson Ettinger \\
Allen Institute for Artificial Intelligence \\
\texttt{allysone@allenai.org }
}

\begin{document}
\maketitle
\begin{abstract}
Selecting high-quality training data can reduce computation cost for LLM fine-tuning. 
Prior work on data selection developed a variety of difficulty-based scores, motivated by the intuition that harder examples are more informative. While effective for smaller classification models, we find that such methods often fail to outperform random selection for LLM fine-tuning. Our analysis reveals that random selection performs better because it is more representative of the underlying input space---i.e., has better coverage.
To validate the importance of coverage, we implement a simple clustering-based selection method which selects data that is more representative of the underlying input distribution. On a set of diverse single-domain tasks, this method yields performance at or above that of random selection. On the diverse, multi-domain T\"{u}lu 3 mixture, where random sampling no longer guarantees a balanced subset, our method outperforms all methods. Overall, our results indicate that prioritizing coverage over difficulty leads to more effective data selection for LLM fine-tuning.
\end{abstract}

\section{Introduction}
\label{sec:introduction}
While scaling---in terms of data, parameters, and compute---has become a standard pathway to success in pre-training LLMs \citep{hoffmann2022trainingcomputeoptimallargelanguage}, it can often conflict with the practicalities (financial or otherwise) of deployment of these models at scale. 
Recent studies have pointed out that, using a small but high-quality dataset can often lead to performance that is competitive with that obtained using a large but potentially noisy dataset \citep{abbas2023semdedupdataefficientlearningwebscale, tirumala2023d}. Similar observations have been made for instruction tuning \citep{zhou2023lima} and fine-tuning for reasoning \citep{ye2025limoreasoning}. 
The field of data selection/pruning, then, presents an alternate pathway to that of data scaling,\footnote{E.g., see \citet{sorscher2022beyond} for how effective data-pruning can allow models to go beyond power-law scaling and achieve exponential scaling} where the goal is to promote sample-efficient training by selecting a high-quality subset of the available data. A promising body of research in data selection has led to the development of various ``difficulty scores''  which leverage the target model's internal signals to rank instances in the training set for level of difficulty. 
As such, difficulty-based data selection has become a popular paradigm in deep-learning, resulting in the proliferation of a variety of difficulty scores \citep[][i.a.]{toneva2018an, swayamdipta-etal-2020-dataset, paul2021deep, agarwal2022estimatingexampledifficultyusing}, primarily applied to classification settings, especially in computer vision tasks. 
While this paradigm has been successful in classification settings, its transfer to LLM fine-tuning is unclear. 
Unlike in standard classification settings, in which the model's output space is constrained to the fixed label-set, LLMs are typically fine-tuned using the same pretraining task of next-word prediction---such that their output space after fine-tuning is still their entire vocabulary.
A major reason to hypothesize that difficulty scores may be insufficient for data selection in LLM fine-tuning is that exclusively prioritizing difficult instances results in only capturing a restricted space of the input data, thus suffering from low data coverage. A specific instance of this argument has previously been explored even in case of classification tasks in computer vision \citep{zheng2023coveragecentriccoresetselectionhigh}, who showed that only including difficult instances leads to catastrophic failures at low data-selection percentages, citing lack of sufficient data coverage as a reason. Similarly, efforts in data filtering for pre-training have highlighted the importance of data diversity, and not difficulty, in order to achieve better performance \citep{tirumala2023d, sachdeva2024traindataefficientllms}. 

Taking these observations as our motivation, in this paper we explore the extent to which difficulty-based data selection transfers to LLM finetuning. We catalog five difficulty scores and extend them to LLM fine-tuning
across three tasks, in which LLMs  have notable gaps between zero-shot and fine-tuning performance, and two models.
Our results from these experiments suggest that transfer of these selection methods for LLM fine-tuning is indeed very weak: random sampling often yields performance that is on par with or sometimes better than these methods. We then hypothesize that this may be due to the abovementioned limitations of difficulty scores in selecting instances with sufficient \textit{coverage} of the training data.
To test this, we first operationalize \textit{coverage} using a method that compares the full data and selected data in terms of distributions over clusters of the data. We find that the divergence between the selected data and the full training data strongly correlates with the performance obtained by fine-tuning LMs on the selected data, supporting our hypothesis. 

Finally, we devise a purely coverage-based selection method that prioritizes data coverage as operationalized in our analysis. We find that while this method outperforms difficulty-based data selection as well as another recently proposed method that balances between difficulty and diversity \citep{maharana2023d2pruningmessagepassing}, it remains on par with random-sampling, only occasionally outperforming it. This finding is consistent with our thesis: if random sampling succeeds because it achieves good coverage, then explicitly optimizing for coverage should converge to random's performance precisely where random already covers the input distribution well. Unlike random sampling, however, our method achieves monotonically increasing performance as more data is selected, avoiding the instability of random selection across selection percentages. 
To further analyze the effect of coverage, we additionally conduct our experiments on T\"{u}lu 3 ~\citep{lambert2025tulu3pushingfrontiers}, a highly diverse SFT mixture used in training \olmo. We find that as the diversity in data increases, coverage-based selection outperforms all other methods at small data percentages.
Overall, our findings support the prioritization of data coverage over difficulty when it comes to LLM fine-tuning.\footnote{https://github.com/LalchandPandia/RethinkingDataSelection}

\section{Related Work}
\paragraph{Data selection using difficulty} 
Difficulty-based methods often use artifacts of models' training dynamics to implicitly or explicitly rank training instances by their ``difficulty''. \citet{swayamdipta-etal-2020-dataset} use model confidence to categorize data instances, while gradient based methods like VoG ~\citep{agarwal2022estimatingexampledifficultyusing, anand-etal-2023-influence} measure difficulty using variance of the gradients. \citet{pmlr-v162-ethayarajh22a} use the notion of usable information for a model to define Pointwise $\mathcal{V}$-usable information (PVI), and use it to quantify the difficulty of data instances.
All of the above methods have been devised for fine-tuning to perform classification tasks. Difficulty-based selection has also been applied to pre-training and alignment task during post-training: ~\citet{marion2023moreinvestigatingdatapruning} use perplexity on a large reference model for data selection during pre-training.
We build on these prior methods by extending difficulty scores for LLM fine-tuning tasks (see \Cref{sec:difficulty-scores}). 

\paragraph{Data selection using diversity}  
Difficulty-based methods often select redundant instances that cover only a limited portion of the data distribution ~\citep{settles-craven-2008-analysis, 10.5555/1757788.1757826}. Prior work in active learning ~\citep{10.1145/1390156.1390183, ash2020deepbatchactivelearning, sener2018active} has 
shown that training on diverse data leads to better performance in computer vision tasks. $\text{D}^2$ ~\citep{maharana2023d2pruningmessagepassing} aims to balance diversity as well as hardness for data selection and is primarily focused on fixed-label classification tasks. The instruction-tuning literature ~\citep{bukharin-etal-2024-data, zhou2023lima} has emphasized the importance of data quality as well as data diversity.  Our focus is to investigate whether diversity alone is sufficient for selecting data when fine-tuning LLMs.

\section{Difficulty Scores}
\label{sec:difficulty-scores}
We approach our experiments and analyses with the goal of identifying scores that will enable effective subselection of data for fine-tuning. We begin by investigating existing difficulty-based scores, which are either derived from the training dynamics of a model trained on the full dataset, or require a single forward pass on the dataset. 
In this section, we discuss our extension of common difficulty scores to LLM fine-tuning.
Common to the data-selection methods used in this work is the notion of a utility function, which determines which instances in the target dataset are selected. Since we are primarily focused on difficulty here, our utility functions will operationalize difficulty in some way, and will prioritize data that are deemed to be more difficult (over those deemed less so).
Given a supervised training dataset ${D_{train} = \{(x_i, y_i)\}_{i=1}^{n}}$, our aim is to choose a subset $S \subset D_{train}$ such that the model trained on $S$ maximizes accuracy on the test set $D_{test}$. Here $y_i$ represents the output sequence instead of a single label.
The utility function $f(x_i): D_{train} \to \mathbb{R}$ maps a data instance $x_i$ to a real number representing the
difficulty score of the instance. 
Once we obtain these difficulty scores, we select $S_{m\%}$, or the subset containing the top-$m$\% of instances ranked by difficulty, as $S_{m\%}=\{x_j \in D_{train} \mid f(x_j) \geq C_{m\%}\}$, where $C_{m\%}$ denotes the $(100-m)$-th percentile of the difficulty score distribution. Below we describe five utility functions used in our analyses.

\paragraph{Perplexity}
Perplexity-based data selection utilizes LMs to evaluate the utility of data points based on model loss for pre-training \citep{marion2023moreinvestigatingdatapruning, ankner2024perplexedperplexityperplexitybaseddata}. 
We adopt perplexity as a difficulty score by selecting instances in the training data with higher perplexity.
\paragraph{Confidence and Variability}
\citet{swayamdipta-etal-2020-dataset} propose two utility functions for classification tasks, both measured across training epochs: (1) a model's confidence of the true label, and (2) its variability. \textbf{Confidence} is calculated as the average likelihood of the true label across epochs, while \textbf{Variability} is its standard deviation. We extend these measures to {fine-tuning LLMs} as follows: let $E$ be the total number of epochs that the LM is fine-tuned for, $x_i$ be the input sequence corresponding to the $i$-th data-instance, and $y_1, \dots, y_n$ be the ground-truth output sequence, then the confidence ($\mu_i$) and variability ($\sigma_i$) are given as:
\begin{equation*}
  \mu_i = \frac{1}{E}\sum_{e=1}^E \left(\prod_{j=1}^{n} p_e(y_j\mid y_{1:j-1}x_i)^{1/n}\right)
\end{equation*}
\begin{equation*}
    \sigma_i = \sqrt{\frac{1}{E}\sum_{e=1}^E \left(\prod_{j=1}^{n} p_e(y_j\mid y_{1:j-1}x_i)^{1/n} - \mu_i\right)^{2}}
\end{equation*}
where $p_e$ is the LM's sequence probability at the $e^{\text{th}}$ epoch. In order to operationalize difficulty-based selection for these metrics, we use smaller values for confidence, and larger values for variability.
\paragraph{Variance of Gradients (VoG)}
VoG \citep{agarwal2022estimatingexampledifficultyusing, anand-etal-2023-influence} is a utility function based on the idea that easy examples lead to saturated losses and stable gradients early in training, whereas ``difficult'' examples exhibit gradient variability throughout training.  We compute the gradients of the logits at the location of target token $y_i$ with respect to the embedding of each word in $x_i$. Averaging the gradients across output length gives us $G_i$. Once $G_i$ is obtained, the computation of VoG follows exactly the same process as the original VoG method. Selection preference is given to instances with high VoG score. For brevity, we describe VoG score calculation in \Cref{sec:vog_extension}
\paragraph{Pointwise $\mathcal{V}$-usable information (PVI)} ~\citet{pmlr-v162-ethayarajh22a} propose PVI, which quantifies the usable information an instance provides to a model for predicting the target. PVI involves fine-tuning two copies of the same base model on  $D_{train}$ and on a modified dataset $ D_{\phi} = \{(\phi, y_i)|(x_i,y_i) \in D_{train}\}$, where all inputs ($x_i$) are replaced with an empty string $\phi$.  
After training, {the method obtains} two models,  $g$ (trained on $D_{train}$), $g'$ (trained on ${D_{\phi}}$).
The PVI for each training instance $(x, y)$ is calculated as
$\text{PVI}(x,y) = -\log_{2}g[\phi](y) + \log_{2}g'[x](y)$.
We apply PVI directly by using the likelihood of the entire output sequence, and select training instances with low PVI as ``difficult'' instances for fine-tuning.
\section{Experimental Setup}\label{sec:experiment_setup}
To test the effectiveness of each of these data selection metrics {in LLM fine-tuning}, we carry out experiments across 3 main datasets, described below.  
For all our experiments, we vary the number of selected instances as \{1, 5, 10, 25, 50, 75\}\% of the total
training data. We first compute our difficulty scores for the entire dataset (i.e., 100\%), use them to select difficult training examples, and then fine-tune a model on the resulting subset.
We compare against a random sampling baseline, in which we randomly sample the given percentage of the training data, and then fine-tune our models on this subset.

\paragraph{Models}
We primarily experiment on two transformer-based language models, both of which are trained on the standard language model objective: Llama 3 8B ~\citep{grattafiori2024llama3herdmodels} and OLMo 2 7B ~\citep{olmo20252olmo2furious}. More experiment details can be found in Appendix ~\ref{sec:hyperparameters}
\paragraph{Datasets}
We choose datasets in which fine-tuning yields marked improvements compared to zero-shot performance for both models (difference of at least 10\% points). Based on this criterion, we select three multiple-choice QA datasets: Social IQa ~\citep{sap-etal-2019-social}, CommonsenseQA~\citep{talmor-etal-2019-commonsenseqa}, and
 CosmosQA ~\citep{huang-etal-2019-cosmos} from among all the datasets that we considered (see \Cref{tab:dataset_considered} for our initial results from five different datasets for both models). We use the validation set of these datasets in our tests.
\paragraph{Evaluation}
We follow the MCF (multiple-choice formulation)-based evaluation procedure \citep{gu-etal-2025-olmes}. That is, we compute accuracy as the exact-match between the first token of the LM's generated response and the true label.
\section{Difficulty-based methods are often outperformed by Random Sampling}
\label{sec:difficulty-result}
\begin{figure*}[!t]
  \includegraphics[width=\textwidth]{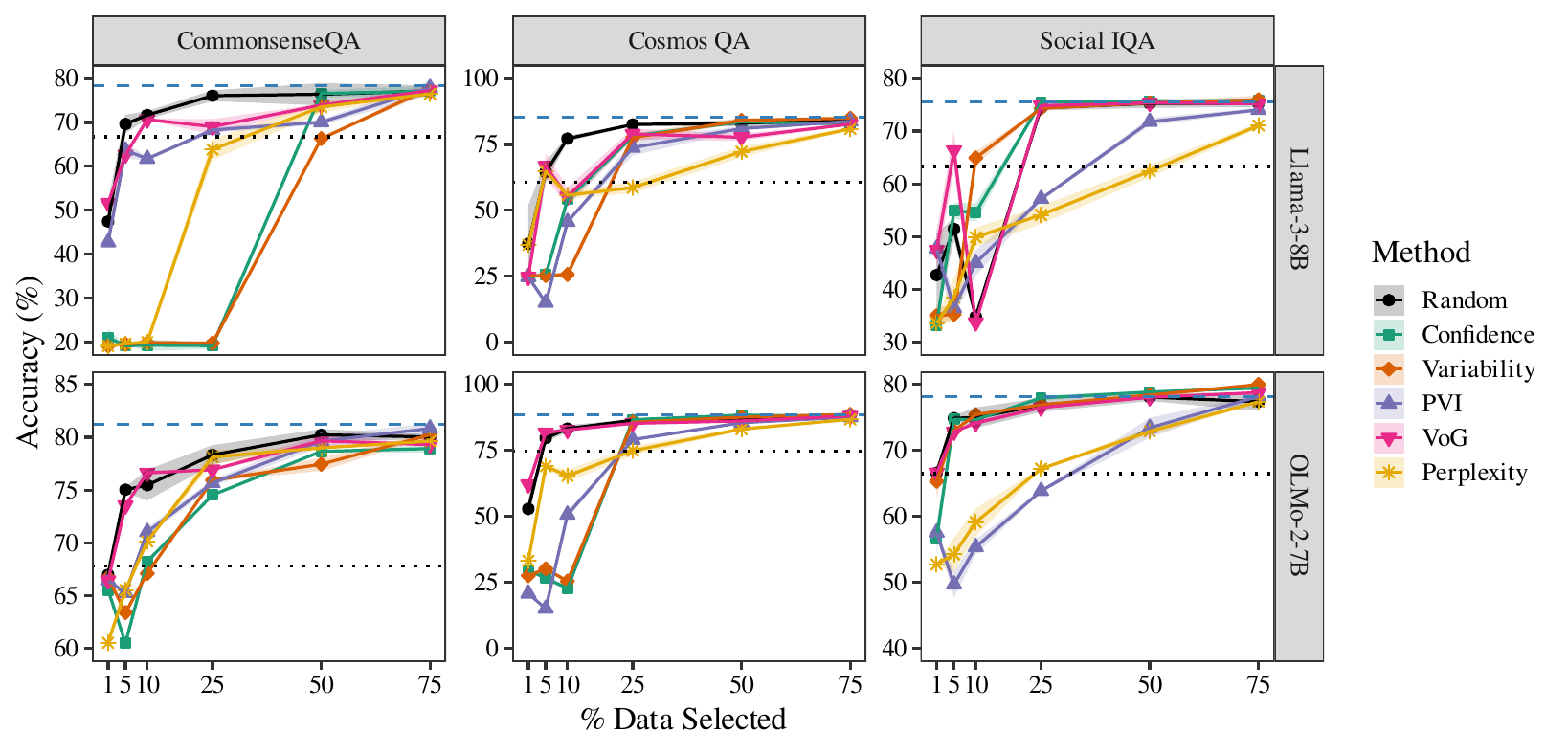}
  \centering
  \caption{Comparative performance of difficulty-based data selection methods on the \llama{} and \olmo{} models across the three datasets, as a function of the percentage of training data selected. For random selection, the points show the mean test accuracy and shaded regions indicate the $\pm1$ standard deviation band across runs. The blue dashed line indicates results from fine-tuning on 100\% training data, while the gray dotted line indicates zero-shot performance. We repeat all our experiments with difficulty-based metrics for 3 random runs where the randomness mainly comes from the training. }
  \label{fig:accuracyWithDifferentScores}
\end{figure*}

We evaluate the difficulty score selection methods as outlined in \Cref{sec:experiment_setup}. This helps us in answering our question: Can we rely on difficulty scores to determine which instances are most useful for fine-tuning LLMs? \Cref{fig:accuracyWithDifferentScores} shows validation accuracies at the end of training as
a function of data selected for each of the five difficulty-based selection methods. 

We first see that models fine-tuned with as little as 50\%
 of the full dataset achieve performance highly comparable to models fine-tuned on the entire dataset. Across all models and datasets, the performance curves for most selection methods---including the random baseline---reach a plateau at or before the 50\% data mark. This suggests that the datasets considered here contain a significant degree of redundancy, and that data selection methods can be highly beneficial for achieving strong performance with reduced training time and computational cost.

We also find that the most effective data selection method
is dependent on the specific dataset but is consistent across models. For example, the variability method consistently yields a top-performing accuracy for the Social IQa dataset on both Llama 3 8B and OLMo 2 7B.
Similarly, for both the CosmosQA and CommonsenseQA datasets,
VoG consistently provides the highest validation accuracy. At an aggregate level, VoG achieves the highest average accuracy among all difficulty-based methods.

Finally, we find that the random baseline is consistently competitive---and often the best performing method---across all selected percentages and datasets. The surprising effectiveness of this simple baseline motivates our subsequent analysis into why random selection outperforms conventional difficulty-based scoring methods.

\section{Difficulty-based selection suffers from poor data coverage}\label{sec:analysis}

Why might simple random sampling be as good as---or even better than---principled, difficulty-based selection methods {in these fine-tuning settings}?
We hypothesize that difficulty-based methods may {select for items occupying a smaller, more restricted} space within the training dataset, {resulting in an unrepresentative sample---whereas a simple random sampling method may cover a larger, more representative space. As a result, a model fine-tuned on a difficulty-based data selection may end up biased toward a smaller space of the {task}, de-prioritizing features that may be important for tasks {that are more general in nature}. This {hypothesis is consistent with the findings of \citet{zheng2023coveragecentriccoresetselectionhigh} for fixed-label computer vision tasks, described in \Cref{sec:introduction}}.
In this section we test this hypothesis by defining a measure of coverage and testing whether this measure is predictive of how effective a selection method will be. To do so, we must first operationalize the notion of \textit{coverage}. However, computing how well a given data-selection scheme picks out a diverse, high-coverage sample of the entire training data requires access to the true distribution of the data, which is intractable. In light of this, we rely on a quantized/discretized embedding space, sensitive to the distributional semantic features
of the instances in the training set. This part of our method is inspired by a similar component introduced in the computation of \textsc{mauve} \citep{pillutla2021mauve}, a metric for comparing LM-generated data to that of humans, using divergence curves.

Our estimation process is as follows: first, we use the non-fine-tuned state of the target LM to map training data instances $\{(x_1, y_1), \dots, (x_n, y_n)\}$ to an embedding space $\textrm{LM}((x_i, y_i)) \rightarrow \mathbb{R}^d$, where $d$ is the dimensionality of the embedding space, giving us a collection of vectors $\{\mathbf{v}_1, \dots, \mathbf{v}_n\}$.\footnote{In practice, we use the model's hidden state representation at the last layer and last position as our embedding extractor.} We then use k-means clustering \citep{lloyd1982least} to quantize these embeddings, allowing us to use the cluster assignments $\{1, \dots, k\}$ as a support to compare the distribution of the training set to a selected subset. To {execute this comparison}, we define $P$ and $Q$,  the (approximate) distributions of the cluster labels in the full training data and a given selected subset, $S_{m\%}$, respectively, as:

\begin{equation*}
    P(j) = \frac{1}{n}\sum_{i=1}^{n}\mathbbm{1}[\psi(\mathbf{v}_i) = j]
\end{equation*}
\begin{equation*}
    Q(j) = \frac{1}{\left|S_{m\%}\right|}\sum_{i=1}^{\left|S_{m\%}\right|}\mathbbm{1}[\psi(\mathbf{v}_i) = j],
\end{equation*}
where $\psi(\mathbf{v})$ represents the cluster assignment function (i.e., one resulting from the k-means algorithm) that returns a cluster label in $\{1, \dots, k\}$. We can now compare how different $Q$ is from $P$, by computing the Jensen-Shannon Divergence (JSD) between them:

\begin{equation*}
    \text{JSD}(P \mid\mid Q) = \frac{1}{2}\text{KL}(P \mid\mid M) + \frac{1}{2}\text{KL}(Q \mid\mid M)
\end{equation*}
\begin{equation*}
    \text{KL}(P \mid\mid Q) = \sum_{x}P(x)\log\frac{P(x)}{Q(x)},
\end{equation*}
where $M$ is the mixture distribution of $P$ and $Q$, computed as $\nicefrac{(P+Q)}{2}$. Since this measure, as it is currently defined, is sensitive to the chosen value of $k$, we repeat this computation over 10 different random seeds, as well as across a range of values for $k$ per dataset, model, and data percentage. More specifically for $m\%$ selected data, we experiment with values of $k$ ranging from 2 to $mn/100$, incrementing in powers of 2.

We now explore how this measure patterns against our findings from the previous section, and whether {it is a reliable predictor of} the effectiveness of different data selection methods.
For this, we measure the average JSD for our data selection methods, across different models and datasets. \Cref{fig:mean-rank} shows the mean rank of each data-selection method, calculated on the basis of its average JSD values across selected-percentages, datasets, and models. We find that the \textbf{Random} data selection method is always the one with the lowest average JSD value across all datasets, both models, and {all} data-selection percentages, suggesting that it systematically achieves the greatest coverage of the training data relative to other methods. Next, we measure the Spearman's correlation between average JSD values of different methods and accuracy obtained as a result of deploying them at each data-selection percentage, across different datasets and models. Insofar as coverage---as operationalized by our metric---is related to data-selection performance, we expect there to be a strong negative correlation between average JSD and accuracy, since methods with poorer coverage (i.e., high divergence) are posited to have poorer performance under this prediction. \Cref{tab:correlation} shows our results across models and datasets. In line with the aforementioned prediction, we find strong negative correspondence between average JSD and accuracy, across all model-dataset pairs, with values ranging from -0.71 to -0.90.

\begin{figure*}[t]
  \begin{minipage}[b]{0.52\textwidth}
    \centering
    \includegraphics[width=\textwidth]{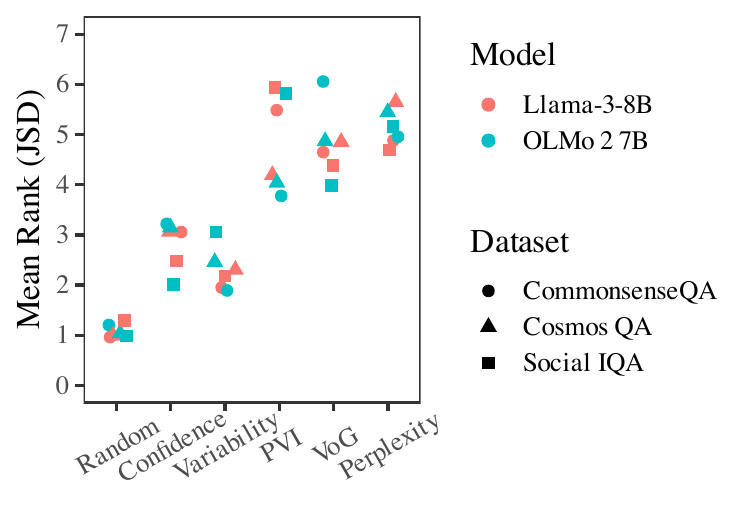}
    \captionof{figure}{Mean rank (based on Avg. JSD) of each selection method across models and datasets. Methods with lower ranks have lower avg. JSD.}
    \label{fig:mean-rank}
  \end{minipage}
  \hfill
  \begin{minipage}[b]{0.45\textwidth}
    \centering
    \begin{tabular}{@{}lcc@{}}
      \toprule
      \textbf{Model} & \textbf{Dataset} & \textbf{$\rho$} \\ \midrule
      \multirow{3}{*}{Llama 3 8B} & CommonsenseQA & -0.71 \\
                                  & CosmosQA      & -0.87 \\
                                  & Social IQa    & -0.81 \\ \midrule
      \multirow{3}{*}{OLMo 2 7B}  & CommonsenseQA & -0.83 \\
                                  & CosmosQA      & -0.88 \\
                                  & Social IQa    & -0.90 \\ \bottomrule
    \end{tabular}
    \captionof{table}{Spearman's correlation between avg. JSD and accuracy across data-selection percentages and methods, for each dataset and model. Lower values indicate greater correspondence between coverage and performance.}
    \label{tab:correlation}
  \end{minipage}
\end{figure*}

Our results suggest that this notion of coverage may explain the weakness of the difficulty scores for selecting subsets of training data: we find that reduced coverage directly correlates with lower task accuracy for models trained on these subsets. Random sampling preserves coverage, {which may be the source of its consistently strong} performance. 
\section{A purely coverage-based method}\label{sec:coverage_based_method}

Our analysis in the previous section indicated that selecting subsets with good coverage of the training data might be key to better LLM fine-tuning performance, and may potentially explain the poor performance of difficulty-based methods. 
To shed further light on the extent to which coverage alone can serve as an alternative to difficulty---and to provide a practical proof of concept for our analysis---we implement a minimal clustering-based selection method. 

Our method builds directly from our analysis in the previous section. We use the embedding space of the training data by using the target LM's last hidden state\footnote{We experiment with different layers as shown in ~\cref{sec:layer_experiment}. However, the last layer gave better results} at the last token-position of each training instance. Then we cluster these embeddings using $k$-means clustering, this time fixing $k$ as $\nicefrac{mn}{100}$, where $n$ is the total size of the training set, and $m$ denotes the data selection percentage.\footnote{We tried different values of $k$ as explained in ~\cref{sec:number_of_clusters}, but this gave the best results}  
We then simply sample one instance from each cluster, allowing all cluster points to be represented, thereby encouraging ``good'' coverage of the data.\footnote{We explored sampling using difficulty measures such as perplexity, etc. but our initial results {in this direction} were {consistently} worse than random sampling.}
We then perform standard fine-tuning as described in \Cref{sec:experiment_setup} to obtain performance estimates that represent our coverage-based selection method. 
We would like to point out that unlike previous works like \citep{tirumala2023d, sorscher2022beyond} we do not rely on pretrained embeddings. We use the target model's embeddings since these may provide a more direct signal of data usefulness.
We compare performance obtained by our method 
to those from random sampling, as well as the best (overall) difficulty-based selection method---i.e., VoG. In addition, we also compare against \textbf{$\text{D}^2$ Pruning} \citep[][abbreviated as $\text{D}^2$]{maharana2023d2pruningmessagepassing}. $\text{D}^2$ aims to balance difficulty and diversity by representing the training set as a graph.
In our adaptation, we use the variability scores from \Cref{sec:difficulty-scores} as node features, following \citet{maharana2023d2pruningmessagepassing}, and the target LLM’s final hidden state at the last token position as embeddings.
Each comparison here meaningfully addresses aspects of our research questions: 1) by comparing against the best difficulty-based method, we directly pit diversity against difficulty; 2) by comparing against $\text{D}^2$ Pruning, we can shed light on how well data-diversity (captured by coverage) alone fares against a selection scheme that balances between the two extremes; and 3) by comparing against random, we test if a more principled and stable method of data-selection can compete against the uncertainty of exclusive random-sampling. 
\begin{figure*}[!t]
  \includegraphics[width=\textwidth]{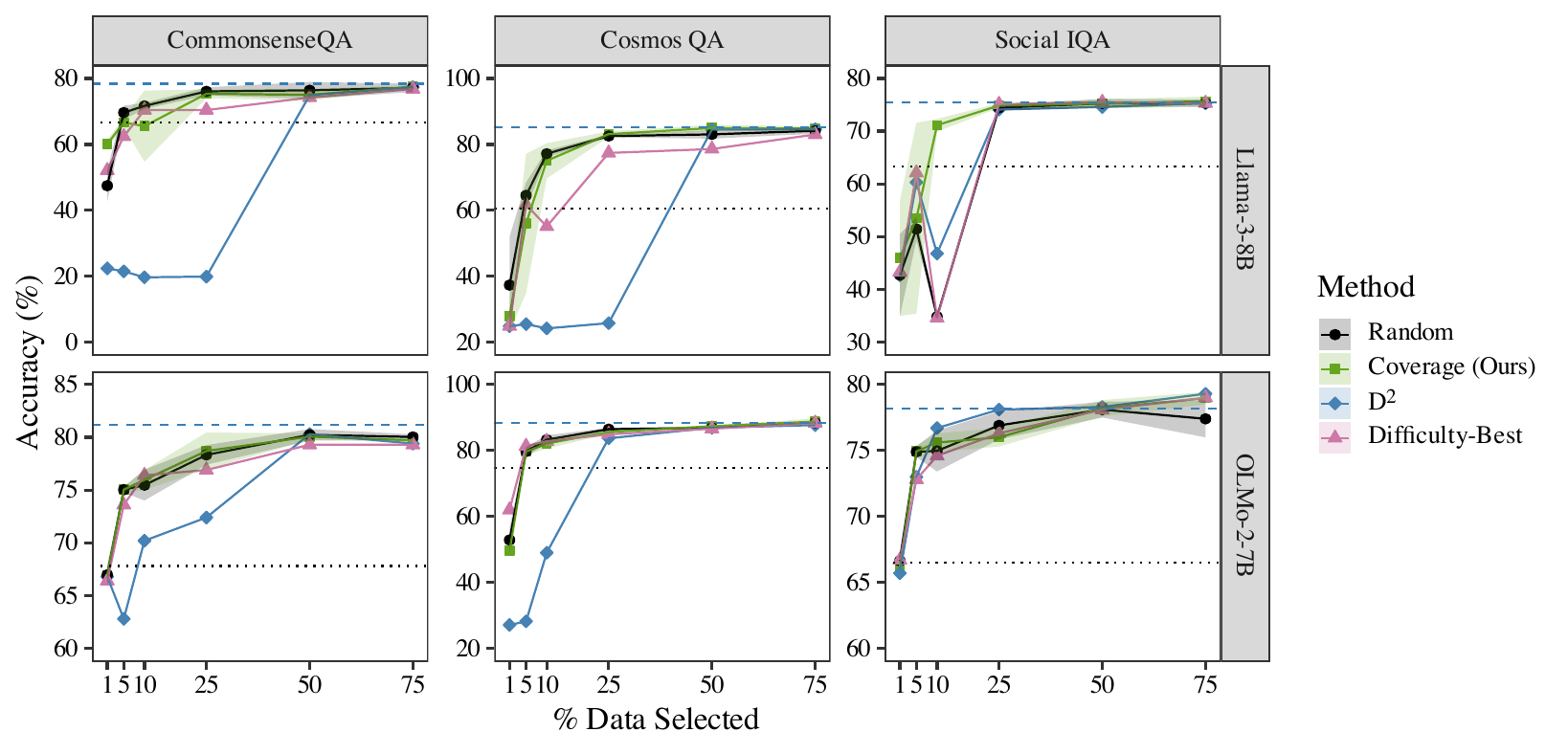}
  \centering
  \caption{Performance of the Coverage-based selection method on Llama 3 8B and OLMo 2 7B models as a function of training data selected, along with comparisons to random selection, the best overall difficulty-based method (VoG), and $\text{D}^2$. For random selection and Coverage, the points show the mean test accuracy and shaded regions indicate the $\pm1$ standard deviation band across runs.~The blue dashed line indicates results from fine-tuning on 100\% training data, while the gray dotted line indicates zero-shot performance.}
  \label{fig:coverage-results}
\end{figure*}
\paragraph{Results} \Cref{fig:coverage-results} {compares} results using our Coverage method to those from random sampling, the best difficulty-based method (VoG), and $\text{D}^2$. First, we see that accuracies obtained from using $\text{D}^2$ are consistently worse at lower percentages, across both models for CommonsenseQA and CosmosQA.
In fact, $\text{D}^2$ seems to under perform even VoG in most cases, suggesting that balancing between difficulty and diversity in this setting harms rather than helps with data selection. There are certain instances where $\text{D}^2$ is in fact the best performing method (e.g., Social IQa)\, but these are overshadowed by {the fact that it performs substantially} worse than zero-shot performance for other datasets.
While VoG is fairly competitive against random and our coverage method, there are cases where it is clearly outperformed by them (e.g., CosmosQA and Social IQa with \llama{}), and it sometimes shows a non-monotonic trajectory with respect to the percentage of data selected. In contrast, our coverage-based method almost always shows an upward trajectory with respect to the selected percentage, obtaining similar performance to the fully fine-tuned model at even 25\% data. Overall, except in a small number of instances, the finding from \Cref{sec:difficulty-result} seems to persist---random sampling is rarely systematically outperformed by any {more principled} method tested so far.
\begin{table*}[t]
\centering
\small
\setlength{\tabcolsep}{4pt}
\begin{tabular}{l ccccc ccccc ccccc}
\toprule
& \multicolumn{5}{c}{\textbf{MMLU} (Full: 62.2)}
& \multicolumn{5}{c}{\textbf{BBH} (Full: 48.5)}
& \multicolumn{5}{c}{\textbf{HumanEval} (Full: 45.9)} \\
\cmidrule(lr){2-6}\cmidrule(lr){7-11}\cmidrule(lr){12-16}
\textbf{Method} & 1\% & 5\% & 10\% & 15\% & 25\%
                & 1\% & 5\% & 10\% & 15\% & 25\%
                & 1\% & 5\% & 10\% & 15\% & 25\% \\
\midrule
VoG             & 61.0 & 59.6 & 60.9 & 61.1 & 61.2 & 45.7 & 46.8 & 45.5 & 47.4 & 48.3 & 30.5 & 33.6 & 32.3 & 30.5 & 32.3 \\
Variability     & 61.6 & 61.8 & 61.8 & 61.6 & 61.9 & \phantom{0}7.5 & \textbf{\underline{49.7}} & \underline{49.4} & \underline{48.8} & 47.9 & 14.9 & 31.5 & 31.9 & 30.9 & 30.6 \\
Perplexity      & 61.6 & 61.0 & 61.6 & 61.5 & 61.7 & 11.5 & 47.2 & \underline{48.7} & \underline{49.0} & 48.3 & \phantom{0}6.5 & 30.3 & 33.1 & 32.6 & 31.4 \\
Confidence      & \textbf{61.7} & 61.6 & 61.5 & 61.3 & 62.1 & 45.2 & \underline{49.5} & \underline{49.1} & 48.0 & 48.0 & 15.7 & 27.3 & 31.3 & 29.9 & 29.5 \\
Random          & 60.9 & 61.2 & 61.6 & 61.9 & 61.7 & 47.4 & \underline{48.8} & \underline{48.5} & \textbf{\underline{49.6}} & 48.2 & \textbf{30.8} & \textbf{36.5} & 31.1 & 36.4 & 37.5 \\
Coverage (Ours) & 61.6 & \textbf{\underline{62.6}} & \textbf{\underline{62.7}} & \textbf{62.1} & \textbf{\underline{62.6}} & \textbf{\underline{48.6}} & \underline{48.8} & \textbf{\underline{49.6}} & \underline{48.9} & \textbf{48.4} & 30.5 & 31.8 & \textbf{34.9} & \textbf{39.6} & \textbf{40.2} \\
\bottomrule
\end{tabular}
\caption{Downstream accuracy (\%) on OLMo 2 7B fine-tuned on subsets of T\"ulu 3, across selection percentages. \textbf{Bold} marks the best method per column; \underline{underline} marks any method that meets or exceeds the full fine-tuning baseline (in header).}
\label{tab:tulu3}
\end{table*}
\subsection{Results on  T\"{u}lu 3}\label{sec:tulu3_experiments}
We additionally validate our findings on T\"{u}lu 3 (downsampled to 50K). Unlike the three QA datasets above, T\"{u}lu 3 spans a wide range of domains, making it a more suitable testbed for our analysis. 
We train \olmo{} using the same setup as described in ~\Cref{sec:experiment_setup}. We evaluate on MMLU~\citep{DBLP:journals/corr/abs-2009-03300}, BBH~\citep{srivastava2023beyond} and HumanEval~\citep{chen2021evaluatinglargelanguagemodels} 
MMLU consists of multiple-choice questions spanning 57 tasks including elementary mathematics, US history, computer science, law, and more; BBH contains challenging reasoning problems for which models benefit from step-by-step reasoning; and  HumanEval evaluates models’ ability to complete Python
code given docstrings. We report 5-shot accuracy for MMLU (averaged across the 57 subtasks), 3-shot exact-match for BBH, and pass@10 (temperature 0.8) for HumanEval, following the answer-extraction procedure of OLMES~\citep{gu-etal-2025-olmes}.

~\Cref{tab:tulu3} shows results at key selection percentages on different downstream benchmarks. 
On a highly diverse multi-domain dataset like T\"{u}lu 3, random sampling is less likely to produce a distributionally balanced subset across domains, and principled coverage-based selection---which actively ensures all regions of the embedding space are represented---becomes especially valuable. On MMLU and BBH, coverage is the only method to consistently match or exceed the full fine-tuning baseline, with as little as 5\% of the data. On HumanEval, coverage's advantage over other selection methods grows with budget suggesting that coverage-based selection generalizes to tasks requiring open-ended generation, including functional code generation evaluated by execution.

\section{Discussion}
Our results suggest that coverage rather than difficulty should be prioritized when doing data selection for LLM {fine-tuning}. This {is indicated by the observation that} random selection and our coverage-based method perform better than difficulty-based scores {for fine-tuning}. The benefit of coverage-based selection appears to scale with dataset diversity: on single-domain tasks random sampling approximates good coverage, but this breaks down on multi-domain data, where T\"{u}lu 3 results show coverage-based method being better.

{Why do we find that lack of coverage leads to such weak performance in fine-tuning?} Diversity has been shown to be a driver of performance during pretraining ~\citep{tirumala2023d, sachdeva2024traindataefficientllms}, {so we postulate that the particular importance we see here for coverage in this setting is attributable to the greater similarity between fine-tuning and more generalized language modeling, by contrast to the more focused decision space of fixed-label classification. More specifically, we speculate that fixed-label classification tasks can be solved by learning a more focused set of items that help to identify the decision boundary, while LLM fine-tuning requires models to retain more generalized capabilities.} 

A natural extension {of this work would be} to combine data difficulty and diversity for data selection.
We {did attempt} to combine difficulty and diversity in our method by selecting data with the highest difficulty (e.g., perplexity) from within each cluster---however, as shown in \Cref{fig:diversity_difficulty}, this approach did not yield any performance gains, suggesting that it was coverage that was driving our observed gains.
\section{Conclusion}
We explored how different difficulty-based data selection methods transfer to LLM fine-tuning. We find that each of these methods fall behind random selection. Our results suggest that the strong performance of random selection arises from its better coverage of the training distribution, {which appears to have greater importance in LLM finetuning}. We also show that a simple {coverage}-based method consistently outperforms difficulty-based methods, and is at par or sometimes better than random sampling. These findings highlight the { importance of coverage in data selection for fine-tuning}. Furthermore, on the diverse T\"{u}lu 3 dataset, coverage-based selection consistently achieves better downstream performance than all other methods. These findings suggest that coverage may be a more important criterion than difficulty for data selection in LLM fine-tuning, with practical importance that grows with dataset diversity.

\section*{Limitations}
Our main goal in this work is to investigate the impact of data coverage and data difficulty while fine-tuning LLMs. While our empirical analyses reveal a strong correlation between coverage and the task performance, establishing causality is left for future work. Furthermore, while notions of difficulty have been shown to fare well in certain pretraining investigations \citep{marion2023moreinvestigatingdatapruning}, they have not been compared to coverage, leaving open the question of difficulty vs. coverage in those settings. 

A practical concern regarding adoption of {any of these} data selection methods is the compute cost for selection. For instance, almost all difficulty-based scores above, except {Perplexity}, require training the model on the full dataset---sometimes even twice (in the case of PVI). This reduces their practical utility in fine-tuning settings, where the cost of selection may offset the efficiency gains of subset training.
By contrast, our coverage-based method requires only a single forward pass over the training data, though this still incurs non-trivial overhead relative to random sampling. Future work can also explore whether selection utility transfers across model scales, as seen in \citet{bordelon2023depthwisehyperparametertransferresidual} for learning rates in Vision Transformers, and in \citet{wang2023oneround} for active learning.
\bibliography{custom}

@article{lloyd1982least,
  title={{Least squares quantization in PCM}},
  author={Lloyd, Stuart},
  journal={IEEE transactions on information theory},
  volume={28},
  number={2},
  pages={129--137},
  year={1982},
  publisher={IEEE}
}

@article{pillutla2021mauve,
  title={Mauve: Measuring the gap between neural text and human text using divergence frontiers},
  author={Pillutla, Krishna and Swayamdipta, Swabha and Zellers, Rowan and Thickstun, John and Welleck, Sean and Choi, Yejin and Harchaoui, Zaid},
  journal={Advances in Neural Information Processing Systems},
  volume={34},
  pages={4816--4828},
  year={2021}
}

@article{sorscher2022beyond,
  title={Beyond neural scaling laws: beating power law scaling via data pruning},
  author={Sorscher, Ben and Geirhos, Robert and Shekhar, Shashank and Ganguli, Surya and Morcos, Ari},
  journal={Advances in Neural Information Processing Systems},
  volume={35},
  pages={19523--19536},
  year={2022}
}

@misc{olmo20252olmo2furious,
      title={2 OLMo 2 Furious}, 
      author={Team OLMo and Pete Walsh and Luca Soldaini and Dirk Groeneveld and Kyle Lo and Shane Arora and Akshita Bhagia and Yuling Gu and Shengyi Huang and Matt Jordan and Nathan Lambert and Dustin Schwenk and Oyvind Tafjord and Taira Anderson and David Atkinson and Faeze Brahman and Christopher Clark and Pradeep Dasigi and Nouha Dziri and Michal Guerquin and Hamish Ivison and Pang Wei Koh and Jiacheng Liu and Saumya Malik and William Merrill and Lester James V. Miranda and Jacob Morrison and Tyler Murray and Crystal Nam and Valentina Pyatkin and Aman Rangapur and Michael Schmitz and Sam Skjonsberg and David Wadden and Christopher Wilhelm and Michael Wilson and Luke Zettlemoyer and Ali Farhadi and Noah A. Smith and Hannaneh Hajishirzi},
      year={2025},
      eprint={2501.00656},
      archivePrefix={arXiv},
      primaryClass={cs.CL},
      url={https://arxiv.org/abs/2501.00656}, 
}

@misc{grattafiori2024llama3herdmodels,
      title={The Llama 3 Herd of Models}, 
      author={Aaron Grattafiori and Abhimanyu Dubey and Abhinav Jauhri and Abhinav Pandey and Abhishek Kadian and Ahmad Al-Dahle and Aiesha Letman and Akhil Mathur and Alan Schelten and Alex Vaughan and Amy Yang and Angela Fan and Anirudh Goyal and Anthony Hartshorn and Aobo Yang and Archi Mitra and Archie Sravankumar and Artem Korenev and Arthur Hinsvark and Arun Rao and Aston Zhang and Aurelien Rodriguez and Austen Gregerson and Ava Spataru and Baptiste Roziere and Bethany Biron and Binh Tang and Bobbie Chern and Charlotte Caucheteux and Chaya Nayak and Chloe Bi and Chris Marra and Chris McConnell and Christian Keller and Christophe Touret and Chunyang Wu and Corinne Wong and Cristian Canton Ferrer and Cyrus Nikolaidis and Damien Allonsius and Daniel Song and Danielle Pintz and Danny Livshits and Danny Wyatt and David Esiobu and Dhruv Choudhary and Dhruv Mahajan and Diego Garcia-Olano and Diego Perino and Dieuwke Hupkes and Egor Lakomkin and Ehab AlBadawy and Elina Lobanova and Emily Dinan and Eric Michael Smith and Filip Radenovic and Francisco Guzmán and Frank Zhang and Gabriel Synnaeve and Gabrielle Lee and Georgia Lewis Anderson and Govind Thattai and Graeme Nail and Gregoire Mialon and Guan Pang and Guillem Cucurell and Hailey Nguyen and Hannah Korevaar and Hu Xu and Hugo Touvron and Iliyan Zarov and Imanol Arrieta Ibarra and Isabel Kloumann and Ishan Misra and Ivan Evtimov and Jack Zhang and Jade Copet and Jaewon Lee and Jan Geffert and Jana Vranes and Jason Park and Jay Mahadeokar and Jeet Shah and Jelmer van der Linde and Jennifer Billock and Jenny Hong and Jenya Lee and Jeremy Fu and Jianfeng Chi and Jianyu Huang and Jiawen Liu and Jie Wang and Jiecao Yu and Joanna Bitton and Joe Spisak and Jongsoo Park and Joseph Rocca and Joshua Johnstun and Joshua Saxe and Junteng Jia and Kalyan Vasuden Alwala and Karthik Prasad and Kartikeya Upasani and Kate Plawiak and Ke Li and Kenneth Heafield and Kevin Stone and Khalid El-Arini and Krithika Iyer and Kshitiz Malik and Kuenley Chiu and Kunal Bhalla and Kushal Lakhotia and Lauren Rantala-Yeary and Laurens van der Maaten and Lawrence Chen and Liang Tan and Liz Jenkins and Louis Martin and Lovish Madaan and Lubo Malo and Lukas Blecher and Lukas Landzaat and Luke de Oliveira and Madeline Muzzi and Mahesh Pasupuleti and Mannat Singh and Manohar Paluri and Marcin Kardas and Maria Tsimpoukelli and Mathew Oldham and Mathieu Rita and Maya Pavlova and Melanie Kambadur and Mike Lewis and Min Si and Mitesh Kumar Singh and Mona Hassan and Naman Goyal and Narjes Torabi and Nikolay Bashlykov and Nikolay Bogoychev and Niladri Chatterji and Ning Zhang and Olivier Duchenne and Onur Çelebi and Patrick Alrassy and Pengchuan Zhang and Pengwei Li and Petar Vasic and Peter Weng and Prajjwal Bhargava and Pratik Dubal and Praveen Krishnan and Punit Singh Koura and Puxin Xu and Qing He and Qingxiao Dong and Ragavan Srinivasan and Raj Ganapathy and Ramon Calderer and Ricardo Silveira Cabral and Robert Stojnic and Roberta Raileanu and Rohan Maheswari and Rohit Girdhar and Rohit Patel and Romain Sauvestre and Ronnie Polidoro and Roshan Sumbaly and Ross Taylor and Ruan Silva and Rui Hou and Rui Wang and Saghar Hosseini and Sahana Chennabasappa and Sanjay Singh and Sean Bell and Seohyun Sonia Kim and Sergey Edunov and Shaoliang Nie and Sharan Narang and Sharath Raparthy and Sheng Shen and Shengye Wan and Shruti Bhosale and Shun Zhang and Simon Vandenhende and Soumya Batra and Spencer Whitman and Sten Sootla and Stephane Collot and Suchin Gururangan and Sydney Borodinsky and Tamar Herman and Tara Fowler and Tarek Sheasha and Thomas Georgiou and Thomas Scialom and Tobias Speckbacher and Todor Mihaylov and Tong Xiao and Ujjwal Karn and Vedanuj Goswami and Vibhor Gupta and Vignesh Ramanathan and Viktor Kerkez and Vincent Gonguet and Virginie Do and Vish Vogeti and Vítor Albiero and Vladan Petrovic and Weiwei Chu and Wenhan Xiong and Wenyin Fu and Whitney Meers and Xavier Martinet and Xiaodong Wang and Xiaofang Wang and Xiaoqing Ellen Tan and Xide Xia and Xinfeng Xie and Xuchao Jia and Xuewei Wang and Yaelle Goldschlag and Yashesh Gaur and Yasmine Babaei and Yi Wen and Yiwen Song and Yuchen Zhang and Yue Li and Yuning Mao and Zacharie Delpierre Coudert and Zheng Yan and Zhengxing Chen and Zoe Papakipos and Aaditya Singh and Aayushi Srivastava and Abha Jain and Adam Kelsey and Adam Shajnfeld and Adithya Gangidi and Adolfo Victoria and Ahuva Goldstand and Ajay Menon and Ajay Sharma and Alex Boesenberg and Alexei Baevski and Allie Feinstein and Amanda Kallet and Amit Sangani and Amos Teo and Anam Yunus and Andrei Lupu and Andres Alvarado and Andrew Caples and Andrew Gu and Andrew Ho and Andrew Poulton and Andrew Ryan and Ankit Ramchandani and Annie Dong and Annie Franco and Anuj Goyal and Aparajita Saraf and Arkabandhu Chowdhury and Ashley Gabriel and Ashwin Bharambe and Assaf Eisenman and Azadeh Yazdan and Beau James and Ben Maurer and Benjamin Leonhardi and Bernie Huang and Beth Loyd and Beto De Paola and Bhargavi Paranjape and Bing Liu and Bo Wu and Boyu Ni and Braden Hancock and Bram Wasti and Brandon Spence and Brani Stojkovic and Brian Gamido and Britt Montalvo and Carl Parker and Carly Burton and Catalina Mejia and Ce Liu and Changhan Wang and Changkyu Kim and Chao Zhou and Chester Hu and Ching-Hsiang Chu and Chris Cai and Chris Tindal and Christoph Feichtenhofer and Cynthia Gao and Damon Civin and Dana Beaty and Daniel Kreymer and Daniel Li and David Adkins and David Xu and Davide Testuggine and Delia David and Devi Parikh and Diana Liskovich and Didem Foss and Dingkang Wang and Duc Le and Dustin Holland and Edward Dowling and Eissa Jamil and Elaine Montgomery and Eleonora Presani and Emily Hahn and Emily Wood and Eric-Tuan Le and Erik Brinkman and Esteban Arcaute and Evan Dunbar and Evan Smothers and Fei Sun and Felix Kreuk and Feng Tian and Filippos Kokkinos and Firat Ozgenel and Francesco Caggioni and Frank Kanayet and Frank Seide and Gabriela Medina Florez and Gabriella Schwarz and Gada Badeer and Georgia Swee and Gil Halpern and Grant Herman and Grigory Sizov and Guangyi and Zhang and Guna Lakshminarayanan and Hakan Inan and Hamid Shojanazeri and Han Zou and Hannah Wang and Hanwen Zha and Haroun Habeeb and Harrison Rudolph and Helen Suk and Henry Aspegren and Hunter Goldman and Hongyuan Zhan and Ibrahim Damlaj and Igor Molybog and Igor Tufanov and Ilias Leontiadis and Irina-Elena Veliche and Itai Gat and Jake Weissman and James Geboski and James Kohli and Janice Lam and Japhet Asher and Jean-Baptiste Gaya and Jeff Marcus and Jeff Tang and Jennifer Chan and Jenny Zhen and Jeremy Reizenstein and Jeremy Teboul and Jessica Zhong and Jian Jin and Jingyi Yang and Joe Cummings and Jon Carvill and Jon Shepard and Jonathan McPhie and Jonathan Torres and Josh Ginsburg and Junjie Wang and Kai Wu and Kam Hou U and Karan Saxena and Kartikay Khandelwal and Katayoun Zand and Kathy Matosich and Kaushik Veeraraghavan and Kelly Michelena and Keqian Li and Kiran Jagadeesh and Kun Huang and Kunal Chawla and Kyle Huang and Lailin Chen and Lakshya Garg and Lavender A and Leandro Silva and Lee Bell and Lei Zhang and Liangpeng Guo and Licheng Yu and Liron Moshkovich and Luca Wehrstedt and Madian Khabsa and Manav Avalani and Manish Bhatt and Martynas Mankus and Matan Hasson and Matthew Lennie and Matthias Reso and Maxim Groshev and Maxim Naumov and Maya Lathi and Meghan Keneally and Miao Liu and Michael L. Seltzer and Michal Valko and Michelle Restrepo and Mihir Patel and Mik Vyatskov and Mikayel Samvelyan and Mike Clark and Mike Macey and Mike Wang and Miquel Jubert Hermoso and Mo Metanat and Mohammad Rastegari and Munish Bansal and Nandhini Santhanam and Natascha Parks and Natasha White and Navyata Bawa and Nayan Singhal and Nick Egebo and Nicolas Usunier and Nikhil Mehta and Nikolay Pavlovich Laptev and Ning Dong and Norman Cheng and Oleg Chernoguz and Olivia Hart and Omkar Salpekar and Ozlem Kalinli and Parkin Kent and Parth Parekh and Paul Saab and Pavan Balaji and Pedro Rittner and Philip Bontrager and Pierre Roux and Piotr Dollar and Polina Zvyagina and Prashant Ratanchandani and Pritish Yuvraj and Qian Liang and Rachad Alao and Rachel Rodriguez and Rafi Ayub and Raghotham Murthy and Raghu Nayani and Rahul Mitra and Rangaprabhu Parthasarathy and Raymond Li and Rebekkah Hogan and Robin Battey and Rocky Wang and Russ Howes and Ruty Rinott and Sachin Mehta and Sachin Siby and Sai Jayesh Bondu and Samyak Datta and Sara Chugh and Sara Hunt and Sargun Dhillon and Sasha Sidorov and Satadru Pan and Saurabh Mahajan and Saurabh Verma and Seiji Yamamoto and Sharadh Ramaswamy and Shaun Lindsay and Shaun Lindsay and Sheng Feng and Shenghao Lin and Shengxin Cindy Zha and Shishir Patil and Shiva Shankar and Shuqiang Zhang and Shuqiang Zhang and Sinong Wang and Sneha Agarwal and Soji Sajuyigbe and Soumith Chintala and Stephanie Max and Stephen Chen and Steve Kehoe and Steve Satterfield and Sudarshan Govindaprasad and Sumit Gupta and Summer Deng and Sungmin Cho and Sunny Virk and Suraj Subramanian and Sy Choudhury and Sydney Goldman and Tal Remez and Tamar Glaser and Tamara Best and Thilo Koehler and Thomas Robinson and Tianhe Li and Tianjun Zhang and Tim Matthews and Timothy Chou and Tzook Shaked and Varun Vontimitta and Victoria Ajayi and Victoria Montanez and Vijai Mohan and Vinay Satish Kumar and Vishal Mangla and Vlad Ionescu and Vlad Poenaru and Vlad Tiberiu Mihailescu and Vladimir Ivanov and Wei Li and Wenchen Wang and Wenwen Jiang and Wes Bouaziz and Will Constable and Xiaocheng Tang and Xiaojian Wu and Xiaolan Wang and Xilun Wu and Xinbo Gao and Yaniv Kleinman and Yanjun Chen and Ye Hu and Ye Jia and Ye Qi and Yenda Li and Yilin Zhang and Ying Zhang and Yossi Adi and Youngjin Nam and Yu and Wang and Yu Zhao and Yuchen Hao and Yundi Qian and Yunlu Li and Yuzi He and Zach Rait and Zachary DeVito and Zef Rosnbrick and Zhaoduo Wen and Zhenyu Yang and Zhiwei Zhao and Zhiyu Ma},
      year={2024},
      eprint={2407.21783},
      archivePrefix={arXiv},
      primaryClass={cs.AI},
      url={https://arxiv.org/abs/2407.21783}, 
}

@misc{hoffmann2022trainingcomputeoptimallargelanguage,
      title={Training Compute-Optimal Large Language Models}, 
      author={Jordan Hoffmann and Sebastian Borgeaud and Arthur Mensch and Elena Buchatskaya and Trevor Cai and Eliza Rutherford and Diego de Las Casas and Lisa Anne Hendricks and Johannes Welbl and Aidan Clark and Tom Hennigan and Eric Noland and Katie Millican and George van den Driessche and Bogdan Damoc and Aurelia Guy and Simon Osindero and Karen Simonyan and Erich Elsen and Jack W. Rae and Oriol Vinyals and Laurent Sifre},
      year={2022},
      eprint={2203.15556},
      archivePrefix={arXiv},
      primaryClass={cs.CL},
      url={https://arxiv.org/abs/2203.15556}, 
}

@misc{abbas2023semdedupdataefficientlearningwebscale,
      title={SemDeDup: Data-efficient learning at web-scale through semantic deduplication}, 
      author={Amro Abbas and Kushal Tirumala and Dániel Simig and Surya Ganguli and Ari S. Morcos},
      year={2023},
      eprint={2303.09540},
      archivePrefix={arXiv},
      primaryClass={cs.LG},
      url={https://arxiv.org/abs/2303.09540}, 
}

@inproceedings{settles-craven-2008-analysis,
    title = "An Analysis of Active Learning Strategies for Sequence Labeling Tasks",
    author = "Settles, Burr  and
      Craven, Mark",
    editor = "Lapata, Mirella  and
      Ng, Hwee Tou",
    booktitle = "Proceedings of the 2008 Conference on Empirical Methods in Natural Language Processing",
    month = oct,
    year = "2008",
    address = "Honolulu, Hawaii",
    publisher = "Association for Computational Linguistics",
    url = "https://aclanthology.org/D08-1112/",
    pages = "1070--1079"
}

@inproceedings{10.5555/1757788.1757826,
author = {Xu, Zhao and Yu, Kai and Tresp, Volker and Xu, Xiaowei and Wang, Jizhi},
title = {Representative sampling for text classification using support vector machines},
year = {2003},
isbn = {3540012745},
publisher = {Springer-Verlag},
address = {Berlin, Heidelberg},
booktitle = {Proceedings of the 25th European Conference on IR Research},
pages = {393–407},
numpages = {15},
location = {Pisa, Italy},
series = {ECIR'03}
}

@inproceedings{
toneva2018an,
title={An Empirical Study of Example Forgetting during Deep Neural Network Learning},
author={Mariya Toneva and Alessandro Sordoni and Remi Tachet des Combes and Adam Trischler and Yoshua Bengio and Geoffrey J. Gordon},
booktitle={International Conference on Learning Representations},
year={2019},
url={https://openreview.net/forum?id=BJlxm30cKm},
}

@misc{bordelon2023depthwisehyperparametertransferresidual,
      title={Depthwise Hyperparameter Transfer in Residual Networks: Dynamics and Scaling Limit}, 
      author={Blake Bordelon and Lorenzo Noci and Mufan Bill Li and Boris Hanin and Cengiz Pehlevan},
      year={2023},
      eprint={2309.16620},
      archivePrefix={arXiv},
      primaryClass={stat.ML},
      url={https://arxiv.org/abs/2309.16620}, 
}

@article{
wang2023oneround,
title={One-Round Active Learning through Data Utility Learning and Proxy Models},
author={Jiachen T. Wang and Si Chen and Ruoxi Jia},
journal={Transactions on Machine Learning Research},
issn={2835-8856},
year={2023},
url={https://openreview.net/forum?id=8HQCOMRa7g},
note={}
}

@misc{sachdeva2024traindataefficientllms,
      title={How to Train Data-Efficient LLMs}, 
      author={Noveen Sachdeva and Benjamin Coleman and Wang-Cheng Kang and Jianmo Ni and Lichan Hong and Ed H. Chi and James Caverlee and Julian McAuley and Derek Zhiyuan Cheng},
      year={2024},
      eprint={2402.09668},
      archivePrefix={arXiv},
      primaryClass={cs.LG},
      url={https://arxiv.org/abs/2402.09668}, 
}

@inproceedings{
tirumala2023d,
title={D4: Improving {LLM} Pretraining via Document De-Duplication and Diversification},
author={Kushal Tirumala and Daniel Simig and Armen Aghajanyan and Ari S. Morcos},
booktitle={Thirty-seventh Conference on Neural Information Processing Systems Datasets and Benchmarks Track},
year={2023},
url={https://openreview.net/forum?id=CG0L2PFrb1}
}

@inproceedings{
paul2021deep,
title={Deep Learning on a Data Diet: Finding Important Examples Early in Training},
author={Mansheej Paul and Surya Ganguli and Gintare Karolina Dziugaite},
booktitle={Advances in Neural Information Processing Systems},
editor={A. Beygelzimer and Y. Dauphin and P. Liang and J. Wortman Vaughan},
year={2021},
url={https://openreview.net/forum?id=Uj7pF-D-YvT}
}

@inproceedings{
zhou2023lima,
title={{LIMA}: Less Is More for Alignment},
author={Chunting Zhou and Pengfei Liu and Puxin Xu and Srini Iyer and Jiao Sun and Yuning Mao and Xuezhe Ma and Avia Efrat and Ping Yu and LILI YU and Susan Zhang and Gargi Ghosh and Mike Lewis and Luke Zettlemoyer and Omer Levy},
booktitle={Thirty-seventh Conference on Neural Information Processing Systems},
year={2023},
url={https://openreview.net/forum?id=KBMOKmX2he}
}

@misc{kaplan2020scalinglawsneurallanguage,
      title={Scaling Laws for Neural Language Models}, 
      author={Jared Kaplan and Sam McCandlish and Tom Henighan and Tom B. Brown and Benjamin Chess and Rewon Child and Scott Gray and Alec Radford and Jeffrey Wu and Dario Amodei},
      year={2020},
      eprint={2001.08361},
      archivePrefix={arXiv},
      primaryClass={cs.LG},
      url={https://arxiv.org/abs/2001.08361}, 
}

@misc{ye2025limoreasoning,
      title={LIMO: Less is More for Reasoning}, 
      author={Yixin Ye and Zhen Huang and Yang Xiao and Ethan Chern and Shijie Xia and Pengfei Liu},
      year={2025},
      eprint={2502.03387},
      archivePrefix={arXiv},
      primaryClass={cs.CL},
      url={https://arxiv.org/abs/2502.03387}, 
}

@inproceedings{swayamdipta-etal-2020-dataset,
    title = "Dataset Cartography: Mapping and Diagnosing Datasets with Training Dynamics",
    author = "Swayamdipta, Swabha  and
      Schwartz, Roy  and
      Lourie, Nicholas  and
      Wang, Yizhong  and
      Hajishirzi, Hannaneh  and
      Smith, Noah A.  and
      Choi, Yejin",
    editor = "Webber, Bonnie  and
      Cohn, Trevor  and
      He, Yulan  and
      Liu, Yang",
    booktitle = "Proceedings of the 2020 Conference on Empirical Methods in Natural Language Processing (EMNLP)",
    month = nov,
    year = "2020",
    address = "Online",
    publisher = "Association for Computational Linguistics",
    url = "https://aclanthology.org/2020.emnlp-main.746",
    doi = "10.18653/v1/2020.emnlp-main.746",
    pages = "9275--9293",
}

@misc{agarwal2022estimatingexampledifficultyusing,
      title={Estimating Example Difficulty Using Variance of Gradients}, 
      author={Chirag Agarwal and Daniel D'souza and Sara Hooker},
      year={2022},
      eprint={2008.11600},
      archivePrefix={arXiv},
      primaryClass={cs.CV},
      url={https://arxiv.org/abs/2008.11600}, 
}

@inproceedings{anand-etal-2023-influence,
    title = "Influence Scores at Scale for Efficient Language Data Sampling",
    author = "Anand, Nikhil  and
      Tan, Joshua  and
      Minakova, Maria",
    editor = "Bouamor, Houda  and
      Pino, Juan  and
      Bali, Kalika",
    booktitle = "Proceedings of the 2023 Conference on Empirical Methods in Natural Language Processing",
    month = dec,
    year = "2023",
    address = "Singapore",
    publisher = "Association for Computational Linguistics",
    url = "https://aclanthology.org/2023.emnlp-main.152/",
    doi = "10.18653/v1/2023.emnlp-main.152",
    pages = "2485--2510"
}

@InProceedings{pmlr-v162-ethayarajh22a,
  title = 	 {Understanding Dataset Difficulty with $\mathcal{V}$-Usable Information},
  author =       {Ethayarajh, Kawin and Choi, Yejin and Swayamdipta, Swabha},
  booktitle = 	 {Proceedings of the 39th International Conference on Machine Learning},
  pages = 	 {5988--6008},
  year = 	 {2022},
  editor = 	 {Chaudhuri, Kamalika and Jegelka, Stefanie and Song, Le and Szepesvari, Csaba and Niu, Gang and Sabato, Sivan},
  volume = 	 {162},
  series = 	 {Proceedings of Machine Learning Research},
  month = 	 {17--23 Jul},
  publisher =    {PMLR},
  url = 	 {https://proceedings.mlr.press/v162/ethayarajh22a.html}
}

@misc{marion2023moreinvestigatingdatapruning,
      title={When Less is More: Investigating Data Pruning for Pretraining LLMs at Scale}, 
      author={Max Marion and Ahmet Üstün and Luiza Pozzobon and Alex Wang and Marzieh Fadaee and Sara Hooker},
      year={2023},
      eprint={2309.04564},
      archivePrefix={arXiv},
      primaryClass={cs.CL},
      url={https://arxiv.org/abs/2309.04564}, 
}

@inproceedings{sap-etal-2019-social,
    title = "Social {IQ}a: Commonsense Reasoning about Social Interactions",
    author = "Sap, Maarten  and
      Rashkin, Hannah  and
      Chen, Derek  and
      Le Bras, Ronan  and
      Choi, Yejin",
    editor = "Inui, Kentaro  and
      Jiang, Jing  and
      Ng, Vincent  and
      Wan, Xiaojun",
    booktitle = "Proceedings of the 2019 Conference on Empirical Methods in Natural Language Processing and the 9th International Joint Conference on Natural Language Processing (EMNLP-IJCNLP)",
    month = nov,
    year = "2019",
    address = "Hong Kong, China",
    publisher = "Association for Computational Linguistics",
    url = "https://aclanthology.org/D19-1454/",
    doi = "10.18653/v1/D19-1454",
    pages = "4463--4473"
}

@inproceedings{talmor-etal-2019-commonsenseqa,
    title = "{C}ommonsense{QA}: A Question Answering Challenge Targeting Commonsense Knowledge",
    author = "Talmor, Alon  and
      Herzig, Jonathan  and
      Lourie, Nicholas  and
      Berant, Jonathan",
    editor = "Burstein, Jill  and
      Doran, Christy  and
      Solorio, Thamar",
    booktitle = "Proceedings of the 2019 Conference of the North {A}merican Chapter of the Association for Computational Linguistics: Human Language Technologies, Volume 1 (Long and Short Papers)",
    month = jun,
    year = "2019",
    address = "Minneapolis, Minnesota",
    publisher = "Association for Computational Linguistics",
    url = "https://aclanthology.org/N19-1421/",
    doi = "10.18653/v1/N19-1421",
    pages = "4149--4158"
}

@inproceedings{huang-etal-2019-cosmos,
    title = "Cosmos {QA}: Machine Reading Comprehension with Contextual Commonsense Reasoning",
    author = "Huang, Lifu  and
      Le Bras, Ronan  and
      Bhagavatula, Chandra  and
      Choi, Yejin",
    editor = "Inui, Kentaro  and
      Jiang, Jing  and
      Ng, Vincent  and
      Wan, Xiaojun",
    booktitle = "Proceedings of the 2019 Conference on Empirical Methods in Natural Language Processing and the 9th International Joint Conference on Natural Language Processing (EMNLP-IJCNLP)",
    month = nov,
    year = "2019",
    address = "Hong Kong, China",
    publisher = "Association for Computational Linguistics",
    url = "https://aclanthology.org/D19-1243/",
    doi = "10.18653/v1/D19-1243",
    pages = "2391--2401"
}

@inproceedings{10.1145/1390156.1390183,
author = {Dasgupta, Sanjoy and Hsu, Daniel},
title = {Hierarchical sampling for active learning},
year = {2008},
isbn = {9781605582054},
publisher = {Association for Computing Machinery},
address = {New York, NY, USA},
url = {https://doi.org/10.1145/1390156.1390183},
doi = {10.1145/1390156.1390183},
booktitle = {Proceedings of the 25th International Conference on Machine Learning},
pages = {208–215},
numpages = {8},
location = {Helsinki, Finland},
series = {ICML '08}
}

@misc{ash2020deepbatchactivelearning,
      title={Deep Batch Active Learning by Diverse, Uncertain Gradient Lower Bounds}, 
      author={Jordan T. Ash and Chicheng Zhang and Akshay Krishnamurthy and John Langford and Alekh Agarwal},
      year={2020},
      eprint={1906.03671},
      archivePrefix={arXiv},
      primaryClass={cs.LG},
      url={https://arxiv.org/abs/1906.03671}, 
}

@misc{zheng2023coveragecentriccoresetselectionhigh,
      title={Coverage-centric Coreset Selection for High Pruning Rates}, 
      author={Haizhong Zheng and Rui Liu and Fan Lai and Atul Prakash},
      year={2023},
      eprint={2210.15809},
      archivePrefix={arXiv},
      primaryClass={cs.LG},
      url={https://arxiv.org/abs/2210.15809}, 
}

@inproceedings{
sener2018active,
title={Active Learning for Convolutional Neural Networks: A Core-Set Approach},
author={Ozan Sener and Silvio Savarese},
booktitle={International Conference on Learning Representations},
year={2018},
url={https://openreview.net/forum?id=H1aIuk-RW},
}

@misc{maharana2023d2pruningmessagepassing,
      title={D2 Pruning: Message Passing for Balancing Diversity and Difficulty in Data Pruning}, 
      author={Adyasha Maharana and Prateek Yadav and Mohit Bansal},
      year={2023},
      eprint={2310.07931},
      archivePrefix={arXiv},
      primaryClass={cs.LG},
      url={https://arxiv.org/abs/2310.07931}, 
}

@inproceedings{DBLP:conf/kdd/RasleyRRH20,
  author={Jeff Rasley and Samyam Rajbhandari and Olatunji Ruwase and Yuxiong He},
  title={DeepSpeed: System Optimizations Enable Training Deep Learning Models with Over 100 Billion Parameters},
  year={2020},
  cdate={1577836800000},
  pages={3505-3506},
  url={https://doi.org/10.1145/3394486.3406703},
  booktitle={KDD},
}

@inproceedings{bukharin-etal-2024-data,
    title = "Data Diversity Matters for Robust Instruction Tuning",
    author = "Bukharin, Alexander  and
      Li, Shiyang  and
      Wang, Zhengyang  and
      Yang, Jingfeng  and
      Yin, Bing  and
      Li, Xian  and
      Zhang, Chao  and
      Zhao, Tuo  and
      Jiang, Haoming",
    editor = "Al-Onaizan, Yaser  and
      Bansal, Mohit  and
      Chen, Yun-Nung",
    booktitle = "Findings of the Association for Computational Linguistics: EMNLP 2024",
    month = nov,
    year = "2024",
    address = "Miami, Florida, USA",
    publisher = "Association for Computational Linguistics",
    url = "https://aclanthology.org/2024.findings-emnlp.195/",
    doi = "10.18653/v1/2024.findings-emnlp.195",
    pages = "3411--3425"
}

@inproceedings{mihaylov-etal-2018-suit,
    title = "Can a Suit of Armor Conduct Electricity? A New Dataset for Open Book Question Answering",
    author = "Mihaylov, Todor  and
      Clark, Peter  and
      Khot, Tushar  and
      Sabharwal, Ashish",
    editor = "Riloff, Ellen  and
      Chiang, David  and
      Hockenmaier, Julia  and
      Tsujii, Jun{'}ichi",
    booktitle = "Proceedings of the 2018 Conference on Empirical Methods in Natural Language Processing",
    month = oct # "-" # nov,
    year = "2018",
    address = "Brussels, Belgium",
    publisher = "Association for Computational Linguistics",
    url = "https://aclanthology.org/D18-1260/",
    doi = "10.18653/v1/D18-1260",
    pages = "2381--2391"
}

@misc{clark2018thinksolvedquestionanswering,
      title={Think you have Solved Question Answering? Try ARC, the AI2 Reasoning Challenge}, 
      author={Peter Clark and Isaac Cowhey and Oren Etzioni and Tushar Khot and Ashish Sabharwal and Carissa Schoenick and Oyvind Tafjord},
      year={2018},
      eprint={1803.05457},
      archivePrefix={arXiv},
      primaryClass={cs.AI},
      url={https://arxiv.org/abs/1803.05457}, 
}

@Misc{accelerate,
  title =        {Accelerate: Training and inference at scale made simple, efficient and adaptable.},
  author =       {Sylvain Gugger and Lysandre Debut and Thomas Wolf and Philipp Schmid and Zachary Mueller and Sourab Mangrulkar and Marc Sun and Benjamin Bossan},
  howpublished = {\url{https://github.com/huggingface/accelerate}},
  year =         {2022}
}

@misc{ankner2024perplexedperplexityperplexitybaseddata,
      title={Perplexed by Perplexity: Perplexity-Based Data Pruning With Small Reference Models}, 
      author={Zachary Ankner and Cody Blakeney and Kartik Sreenivasan and Max Marion and Matthew L. Leavitt and Mansheej Paul},
      year={2024},
      eprint={2405.20541},
      archivePrefix={arXiv},
      primaryClass={cs.LG},
      url={https://arxiv.org/abs/2405.20541}, 
}

@misc{lambert2025tulu3pushingfrontiers,
      title={Tulu 3: Pushing Frontiers in Open Language Model Post-Training}, 
      author={Nathan Lambert and Jacob Morrison and Valentina Pyatkin and Shengyi Huang and Hamish Ivison and Faeze Brahman and Lester James V. Miranda and Alisa Liu and Nouha Dziri and Shane Lyu and Yuling Gu and Saumya Malik and Victoria Graf and Jena D. Hwang and Jiangjiang Yang and Ronan Le Bras and Oyvind Tafjord and Chris Wilhelm and Luca Soldaini and Noah A. Smith and Yizhong Wang and Pradeep Dasigi and Hannaneh Hajishirzi},
      year={2025},
      eprint={2411.15124},
      archivePrefix={arXiv},
      primaryClass={cs.CL},
      url={https://arxiv.org/abs/2411.15124}, 
}

@inproceedings{gu-etal-2025-olmes,
    title = "{OLMES}: A Standard for Language Model Evaluations",
    author = "Gu, Yuling  and
      Tafjord, Oyvind  and
      Kuehl, Bailey  and
      Haddad, Dany  and
      Dodge, Jesse  and
      Hajishirzi, Hannaneh",
    editor = "Chiruzzo, Luis  and
      Ritter, Alan  and
      Wang, Lu",
    booktitle = "Findings of the Association for Computational Linguistics: NAACL 2025",
    month = apr,
    year = "2025",
    address = "Albuquerque, New Mexico",
    publisher = "Association for Computational Linguistics",
    url = "https://aclanthology.org/2025.findings-naacl.282/",
    doi = "10.18653/v1/2025.findings-naacl.282",
    pages = "5005--5033",
    ISBN = "979-8-89176-195-7"
}

@misc{chen2021evaluatinglargelanguagemodels,
      title={Evaluating Large Language Models Trained on Code}, 
      author={Mark Chen and Jerry Tworek and Heewoo Jun and Qiming Yuan and Henrique Ponde de Oliveira Pinto and Jared Kaplan and Harri Edwards and Yuri Burda and Nicholas Joseph and Greg Brockman and Alex Ray and Raul Puri and Gretchen Krueger and Michael Petrov and Heidy Khlaaf and Girish Sastry and Pamela Mishkin and Brooke Chan and Scott Gray and Nick Ryder and Mikhail Pavlov and Alethea Power and Lukasz Kaiser and Mohammad Bavarian and Clemens Winter and Philippe Tillet and Felipe Petroski Such and Dave Cummings and Matthias Plappert and Fotios Chantzis and Elizabeth Barnes and Ariel Herbert-Voss and William Hebgen Guss and Alex Nichol and Alex Paino and Nikolas Tezak and Jie Tang and Igor Babuschkin and Suchir Balaji and Shantanu Jain and William Saunders and Christopher Hesse and Andrew N. Carr and Jan Leike and Josh Achiam and Vedant Misra and Evan Morikawa and Alec Radford and Matthew Knight and Miles Brundage and Mira Murati and Katie Mayer and Peter Welinder and Bob McGrew and Dario Amodei and Sam McCandlish and Ilya Sutskever and Wojciech Zaremba},
      year={2021},
      eprint={2107.03374},
      archivePrefix={arXiv},
      primaryClass={cs.LG},
      url={https://arxiv.org/abs/2107.03374}, 
}

@article{
srivastava2023beyond,
title={Beyond the Imitation Game: Quantifying and extrapolating the capabilities of language models},
author={Aarohi Srivastava and Abhinav Rastogi and Abhishek Rao and Abu Awal Md Shoeb and Abubakar Abid and Adam Fisch and Adam R. Brown and Adam Santoro and Aditya Gupta and Adri{\`a} Garriga-Alonso and Agnieszka Kluska and Aitor Lewkowycz and Akshat Agarwal and Alethea Power and Alex Ray and Alex Warstadt and Alexander W. Kocurek and Ali Safaya and Ali Tazarv and Alice Xiang and Alicia Parrish and Allen Nie and Aman Hussain and Amanda Askell and Amanda Dsouza and Ambrose Slone and Ameet Rahane and Anantharaman S. Iyer and Anders Johan Andreassen and Andrea Madotto and Andrea Santilli and Andreas Stuhlm{\"u}ller and Andrew M. Dai and Andrew La and Andrew Kyle Lampinen and Andy Zou and Angela Jiang and Angelica Chen and Anh Vuong and Animesh Gupta and Anna Gottardi and Antonio Norelli and Anu Venkatesh and Arash Gholamidavoodi and Arfa Tabassum and Arul Menezes and Arun Kirubarajan and Asher Mullokandov and Ashish Sabharwal and Austin Herrick and Avia Efrat and Aykut Erdem and Ayla Karaka{\c{s}} and B. Ryan Roberts and Bao Sheng Loe and Barret Zoph and Bart{\l}omiej Bojanowski and Batuhan {\"O}zyurt and Behnam Hedayatnia and Behnam Neyshabur and Benjamin Inden and Benno Stein and Berk Ekmekci and Bill Yuchen Lin and Blake Howald and Bryan Orinion and Cameron Diao and Cameron Dour and Catherine Stinson and Cedrick Argueta and Cesar Ferri and Chandan Singh and Charles Rathkopf and Chenlin Meng and Chitta Baral and Chiyu Wu and Chris Callison-Burch and Christopher Waites and Christian Voigt and Christopher D Manning and Christopher Potts and Cindy Ramirez and Clara E. Rivera and Clemencia Siro and Colin Raffel and Courtney Ashcraft and Cristina Garbacea and Damien Sileo and Dan Garrette and Dan Hendrycks and Dan Kilman and Dan Roth and C. Daniel Freeman and Daniel Khashabi and Daniel Levy and Daniel Mosegu{\'\i} Gonz{\'a}lez and Danielle Perszyk and Danny Hernandez and Danqi Chen and Daphne Ippolito and Dar Gilboa and David Dohan and David Drakard and David Jurgens and Debajyoti Datta and Deep Ganguli and Denis Emelin and Denis Kleyko and Deniz Yuret and Derek Chen and Derek Tam and Dieuwke Hupkes and Diganta Misra and Dilyar Buzan and Dimitri Coelho Mollo and Diyi Yang and Dong-Ho Lee and Dylan Schrader and Ekaterina Shutova and Ekin Dogus Cubuk and Elad Segal and Eleanor Hagerman and Elizabeth Barnes and Elizabeth Donoway and Ellie Pavlick and Emanuele Rodol{\`a} and Emma Lam and Eric Chu and Eric Tang and Erkut Erdem and Ernie Chang and Ethan A Chi and Ethan Dyer and Ethan Jerzak and Ethan Kim and Eunice Engefu Manyasi and Evgenii Zheltonozhskii and Fanyue Xia and Fatemeh Siar and Fernando Mart{\'\i}nez-Plumed and Francesca Happ{\'e} and Francois Chollet and Frieda Rong and Gaurav Mishra and Genta Indra Winata and Gerard de Melo and Germ{\`a}n Kruszewski and Giambattista Parascandolo and Giorgio Mariani and Gloria Xinyue Wang and Gonzalo Jaimovitch-Lopez and Gregor Betz and Guy Gur-Ari and Hana Galijasevic and Hannah Kim and Hannah Rashkin and Hannaneh Hajishirzi and Harsh Mehta and Hayden Bogar and Henry Francis Anthony Shevlin and Hinrich Schuetze and Hiromu Yakura and Hongming Zhang and Hugh Mee Wong and Ian Ng and Isaac Noble and Jaap Jumelet and Jack Geissinger and Jackson Kernion and Jacob Hilton and Jaehoon Lee and Jaime Fern{\'a}ndez Fisac and James B Simon and James Koppel and James Zheng and James Zou and Jan Kocon and Jana Thompson and Janelle Wingfield and Jared Kaplan and Jarema Radom and Jascha Sohl-Dickstein and Jason Phang and Jason Wei and Jason Yosinski and Jekaterina Novikova and Jelle Bosscher and Jennifer Marsh and Jeremy Kim and Jeroen Taal and Jesse Engel and Jesujoba Alabi and Jiacheng Xu and Jiaming Song and Jillian Tang and Joan Waweru and John Burden and John Miller and John U. Balis and Jonathan Batchelder and Jonathan Berant and J{\"o}rg Frohberg and Jos Rozen and Jose Hernandez-Orallo and Joseph Boudeman and Joseph Guerr and Joseph Jones and Joshua B. Tenenbaum and Joshua S. Rule and Joyce Chua and Kamil Kanclerz and Karen Livescu and Karl Krauth and Karthik Gopalakrishnan and Katerina Ignatyeva and Katja Markert and Kaustubh Dhole and Kevin Gimpel and Kevin Omondi and Kory Wallace Mathewson and Kristen Chiafullo and Ksenia Shkaruta and Kumar Shridhar and Kyle McDonell and Kyle Richardson and Laria Reynolds and Leo Gao and Li Zhang and Liam Dugan and Lianhui Qin and Lidia Contreras-Ochando and Louis-Philippe Morency and Luca Moschella and Lucas Lam and Lucy Noble and Ludwig Schmidt and Luheng He and Luis Oliveros-Col{\'o}n and Luke Metz and L{\"u}tfi Kerem Senel and Maarten Bosma and Maarten Sap and Maartje Ter Hoeve and Maheen Farooqi and Manaal Faruqui and Mantas Mazeika and Marco Baturan and Marco Marelli and Marco Maru and Maria Jose Ramirez-Quintana and Marie Tolkiehn and Mario Giulianelli and Martha Lewis and Martin Potthast and Matthew L Leavitt and Matthias Hagen and M{\'a}ty{\'a}s Schubert and Medina Orduna Baitemirova and Melody Arnaud and Melvin McElrath and Michael Andrew Yee and Michael Cohen and Michael Gu and Michael Ivanitskiy and Michael Starritt and Michael Strube and Micha{\l} Sw{\k{e}}drowski and Michele Bevilacqua and Michihiro Yasunaga and Mihir Kale and Mike Cain and Mimee Xu and Mirac Suzgun and Mitch Walker and Mo Tiwari and Mohit Bansal and Moin Aminnaseri and Mor Geva and Mozhdeh Gheini and Mukund Varma T and Nanyun Peng and Nathan Andrew Chi and Nayeon Lee and Neta Gur-Ari Krakover and Nicholas Cameron and Nicholas Roberts and Nick Doiron and Nicole Martinez and Nikita Nangia and Niklas Deckers and Niklas Muennighoff and Nitish Shirish Keskar and Niveditha S. Iyer and Noah Constant and Noah Fiedel and Nuan Wen and Oliver Zhang and Omar Agha and Omar Elbaghdadi and Omer Levy and Owain Evans and Pablo Antonio Moreno Casares and Parth Doshi and Pascale Fung and Paul Pu Liang and Paul Vicol and Pegah Alipoormolabashi and Peiyuan Liao and Percy Liang and Peter W Chang and Peter Eckersley and Phu Mon Htut and Pinyu Hwang and Piotr Mi{\l}kowski and Piyush Patil and Pouya Pezeshkpour and Priti Oli and Qiaozhu Mei and Qing Lyu and Qinlang Chen and Rabin Banjade and Rachel Etta Rudolph and Raefer Gabriel and Rahel Habacker and Ramon Risco and Rapha{\"e}l Milli{\`e}re and Rhythm Garg and Richard Barnes and Rif A. Saurous and Riku Arakawa and Robbe Raymaekers and Robert Frank and Rohan Sikand and Roman Novak and Roman Sitelew and Ronan Le Bras and Rosanne Liu and Rowan Jacobs and Rui Zhang and Russ Salakhutdinov and Ryan Andrew Chi and Seungjae Ryan Lee and Ryan Stovall and Ryan Teehan and Rylan Yang and Sahib Singh and Saif M. Mohammad and Sajant Anand and Sam Dillavou and Sam Shleifer and Sam Wiseman and Samuel Gruetter and Samuel R. Bowman and Samuel Stern Schoenholz and Sanghyun Han and Sanjeev Kwatra and Sarah A. Rous and Sarik Ghazarian and Sayan Ghosh and Sean Casey and Sebastian Bischoff and Sebastian Gehrmann and Sebastian Schuster and Sepideh Sadeghi and Shadi Hamdan and Sharon Zhou and Shashank Srivastava and Sherry Shi and Shikhar Singh and Shima Asaadi and Shixiang Shane Gu and Shubh Pachchigar and Shubham Toshniwal and Shyam Upadhyay and Shyamolima Shammie Debnath and Siamak Shakeri and Simon Thormeyer and Simone Melzi and Siva Reddy and Sneha Priscilla Makini and Soo-Hwan Lee and Spencer Torene and Sriharsha Hatwar and Stanislas Dehaene and Stefan Divic and Stefano Ermon and Stella Biderman and Stephanie Lin and Stephen Prasad and Steven Piantadosi and Stuart Shieber and Summer Misherghi and Svetlana Kiritchenko and Swaroop Mishra and Tal Linzen and Tal Schuster and Tao Li and Tao Yu and Tariq Ali and Tatsunori Hashimoto and Te-Lin Wu and Th{\'e}o Desbordes and Theodore Rothschild and Thomas Phan and Tianle Wang and Tiberius Nkinyili and Timo Schick and Timofei Kornev and Titus Tunduny and Tobias Gerstenberg and Trenton Chang and Trishala Neeraj and Tushar Khot and Tyler Shultz and Uri Shaham and Vedant Misra and Vera Demberg and Victoria Nyamai and Vikas Raunak and Vinay Venkatesh Ramasesh and vinay uday prabhu and Vishakh Padmakumar and Vivek Srikumar and William Fedus and William Saunders and William Zhang and Wout Vossen and Xiang Ren and Xiaoyu Tong and Xinran Zhao and Xinyi Wu and Xudong Shen and Yadollah Yaghoobzadeh and Yair Lakretz and Yangqiu Song and Yasaman Bahri and Yejin Choi and Yichi Yang and Sophie Hao and Yifu Chen and Yonatan Belinkov and Yu Hou and Yufang Hou and Yuntao Bai and Zachary Seid and Zhuoye Zhao and Zijian Wang and Zijie J. Wang and Zirui Wang and Ziyi Wu},
journal={Transactions on Machine Learning Research},
issn={2835-8856},
year={2023},
url={https://openreview.net/forum?id=uyTL5Bvosj},
note={Featured Certification}
}

@article{DBLP:journals/corr/abs-2009-03300,
  author       = {Dan Hendrycks and
                  Collin Burns and
                  Steven Basart and
                  Andy Zou and
                  Mantas Mazeika and
                  Dawn Song and
                  Jacob Steinhardt},
  title        = {Measuring Massive Multitask Language Understanding},
  journal      = {CoRR},
  volume       = {abs/2009.03300},
  year         = {2020},
  url          = {https://arxiv.org/abs/2009.03300},
  eprinttype   = {arXiv},
  eprint       = {2009.03300},
  bibsource    = {dblp computer science bibliography, https://dblp.org}
}

\appendix

\section{Appendix}
\label{sec:appendix}
\subsection{Effect of Random Selection vs. Perplexity in our Coverage Method}

\Cref{fig:diversity_difficulty} shows the impact of using perplexity vs. random sampling for our Coverage method. Using perplexity does not provide any noticeable gain, and in fact hurts performance at lower data-percentages---e.g., see performance on \llama{} for CommonsenseQA, or on \olmo{} for Social IQa.

\begin{figure*}[ht]
  \includegraphics[width=\textwidth]{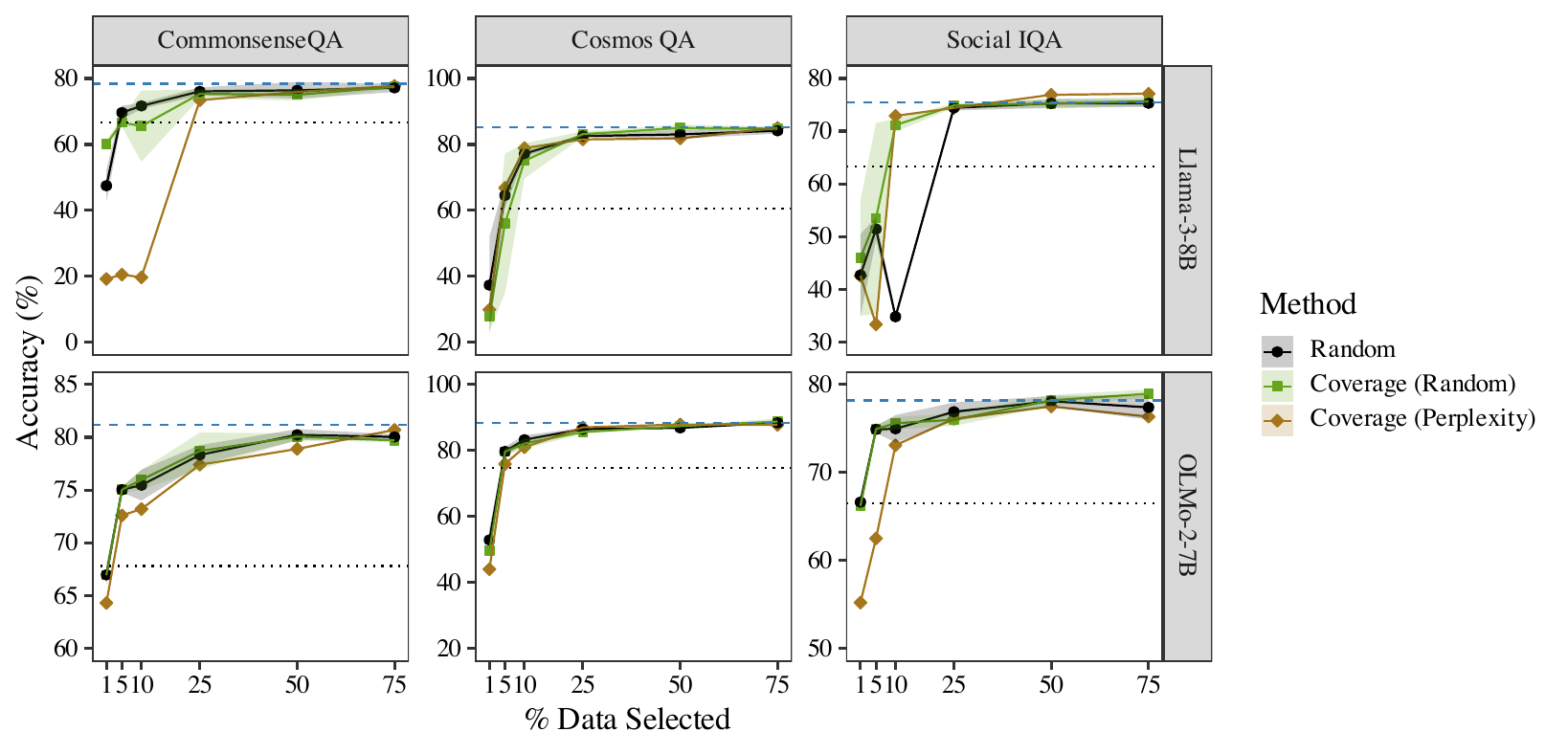}
  \centering
  \caption{Effect of random sampling vs. selection using perplexity after performing clustering for our coverage-based method.}
  \label{fig:diversity_difficulty}
\end{figure*}

\begin{table*}[ht]
\centering
\begin{tabular}{p{.2\textwidth}|p{.2\textwidth}|p{.2\textwidth}|p{.2\textwidth}}
\toprule
\textbf{Dataset} & \textbf{Model} &\textbf{Zero-shot} & \textbf{100\%}\\
\midrule
Social IQa&Llama 3 8B&0.633&0.755\\
&OLMo 2 7B&0.665&0.782\\
\midrule
CommonsenseQA&Llama 3 8B&0.667&0.784\\
&OLMo 2 7B&0.678&0.812\\
\midrule
CosmosQA&Llama 3&0.606&0.853 \\
&OLMo 2 7B&0.747& 0.884\\
\midrule
OpenBookQA~\citep{mihaylov-etal-2018-suit}&Llama 3 8B&0.736&0.828\\
&OLMo 2 7B&0.714&0.844\\
\midrule
ARC-Hard~\citep{clark2018thinksolvedquestionanswering}&Llama 3 8B&0.782& 0.743\\
&OLMo 2 7B&0.709& 0.799\\
\bottomrule
\end{tabular}
\caption{Datasets considered for experiments. Zero-shot performance and performance on using the entire train data on Llama 3 and OLMo 2}\label{tab:dataset_considered}
\end{table*}
\subsection{VoG Extension}\label{sec:vog_extension}
Let $x_i = x_{i,1},x_{i,2},...x_{i,N}$ and $y_i = y_{i,1},y_{i,2},...y_{i,M}$. Let $A_{i,j}$ be the logit at the location of $j-th$ token in $y_i$. Now, $G_{i,k} = \frac{\sum_{j=1}^{M}\frac{\partial A_{i,j}}{\partial E_{x_{i,k}}}}{M}$. \\
Here, $G_{i,k}$ represents the gradient of logit at the target location $y_j$ w.r.t the embedding vector of $k-th$ token in $x_i$.
We obtain $G_i = G_{i,1}|G_{i,2}|...G_{i,N}$ by concatenating the obtained logit gradients. We calculate the mean and variance  across the epochs.
$\mu_i = \frac{\sum_{e=1}^{E}G_i}{E}$\\
$V_i = \frac{{(G_i^{e} - \mu_i})^2}{\sqrt{E}}$\\
The (unnormalized) score $v_i$ for each instance $x_i$ is
then given by the mean of $V_i$ (that is, we average over the input embeddings)\\
$VoG(x_i) =\frac{v_i - \mu_{dset}}{\sigma_{dset}}$\\
Here, $\mu_{dset}$ and $\sigma_{dset}$ are the mean and standard deviation of $v_i$ for the full dataset.
\subsection{Data Selection FLOPS}\label{sec:flop_cost}
For FLOP calculation, we use the common approximation as performed by ~\citet{kaplan2020scalinglawsneurallanguage} $C_{forward} = 2ND$ for forward pass and $C_{train}=6ND$ for one finetuning pass  which involves one forward and one backward pass, where $D$ denotes total number of training
tokens, and $N$ is the number of parameters. \Cref{fig:flop-accuracy} shows a comparison between methods across models and datasets in terms of their average accuracy vs. FLOPs used during data selection.

\begin{table*}[ht]
\centering
\begin{tabular}{p{.2\textwidth}|p{.4\textwidth}|p{.4\textwidth}}
\toprule
\textbf{Dataset} & \textbf{Model} & \textbf{Total tokens}\\
\midrule
Social IQa & Llama 3 8B &1,786,531\\

 & OLMo 2 7B &1,786,578\\
\midrule
CommonsenseQA & Llama 3 8B &577,187\\

& OLMo 2 7B &577,230\\

\midrule
CosmosQA & Llama 3 8B &3,692,490\\

  & OLMo 2 7B &3,692,921\\
 \bottomrule
\end{tabular}
\caption{Total tokens.}\label{tab:flops}
\end{table*}

\begin{figure*}
    \centering
    \includegraphics[width=\textwidth]{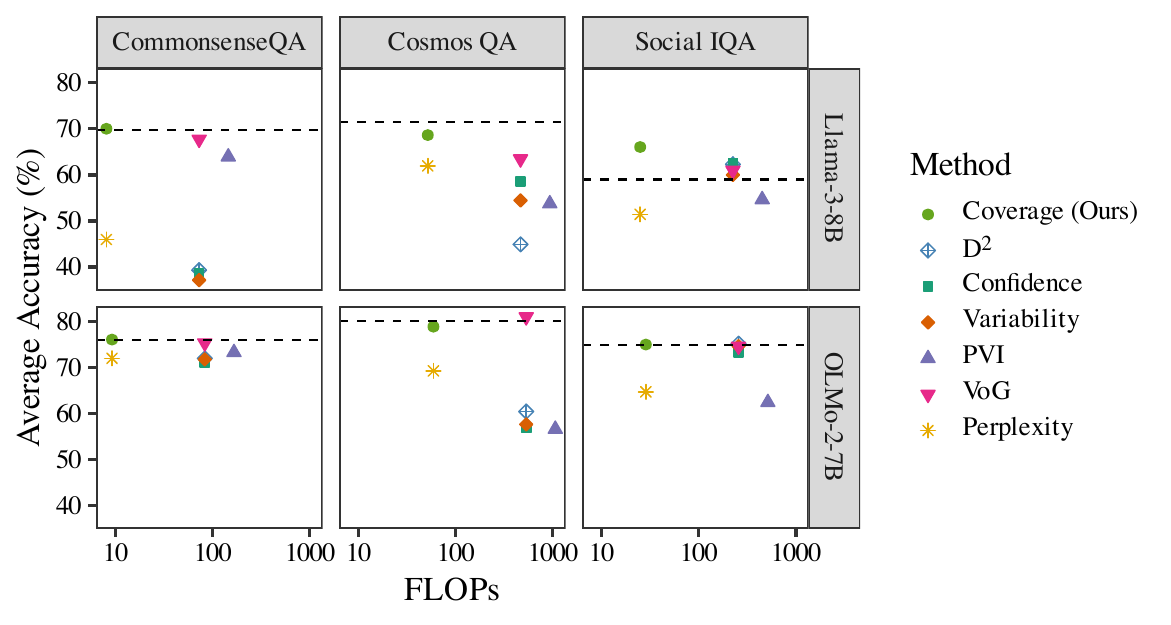}
    \caption{Data Selection FLOPs ($\times 10^{15}$) vs. Average Accuracy (calculated over all percentages) across all methods. X axis is in logarithmic scale. Dashed line indicates Random Selection performance which has a data-selection FLOP value of 0.}
    \label{fig:flop-accuracy}
\end{figure*}

\begin{table*}[ht]
\centering  
\begin{tabular}{p{.2\textwidth} p{.4\textwidth} p{.4\textwidth}}
\toprule
\textbf{Difficulty Score} & \textbf{Dataset} & \textbf{Data Selection FLOPs ($\times 10^{15}$)} \\
\midrule
Perplexity & Social IQa & 25\\
 & CommonsenseQA & 8.08\\
 & CosmosQA & 51.7\\
  \midrule
  Confidence & Social IQa & 225\\
 & CommonsenseQA & 72.7\\
 & CosmosQA & 465\\
   \midrule
  Variability & Social IQa & 225\\
 & CommonsenseQA & 72.7\\
 & CosmosQA & 465\\

\midrule
PVI & Social IQa & 450\\
 & CommonsenseQA & 145\\
 & CosmosQA & 931\\
\midrule
VoG & Social IQa & 225\\
 & CommonsenseQA & 72.7\\
 & CosmosQA &465 \\

 \bottomrule
\end{tabular}
\caption{FLOPS utilized for data selection using Llama 3 8B.}\label{tab:flops_llama}
\end{table*}

\begin{table*}[ht]
\centering
\begin{tabular}{p{.2\textwidth}|p{.4\textwidth}|p{.4\textwidth}}
\toprule
\textbf{Difficulty Score} & \textbf{Dataset} & \textbf{Data Selection FLOPs ($\times 10^{15}$)} \\
\midrule
Perplexity & Social IQa & 28.6\\
 & CommonsenseQA & 9.23\\
 & CosmosQA & 59.1\\
  \midrule
  Confidence & Social IQa &  257\\
 & CommonsenseQA & 83.1\\
 & CosmosQA & 532\\
   \midrule
  Variability & Social IQa &  257\\
 & CommonsenseQA & 83.1\\
 & CosmosQA & 532\\

\midrule
PVI & Social IQa & 515\\
 & CommonsenseQA & 166\\
 & CosmosQA &1060 \\
\midrule
VoG & Social IQa &  257\\
 & CommonsenseQA & 83.1\\
 & CosmosQA &532 \\

 \bottomrule
\end{tabular}
\caption{FLOPS utilized for data selection using OLMo 2 7B.}\label{tab:flops_olmo}
\end{table*}
\subsection{Impact of number of cluster on our method}\label{sec:number_of_clusters}
In our method, we carry out clustering step over embeddings. In the main body of the paper, we utilize the heuristic $K = \frac{mn}{100}$  for all experiments. To validate this choice and assess the stability of our method, we conducted an ablation study varying $K$ across a relevant range when $m$ = 25. We tested values of $K$ values which are powers of two closest to and below the proposed heuristic $K=\frac{mn}{100}$. We evaluate the accuracy, across these variations.  As can be seen in ~\Cref{fig:accuracyWithDifferentK}, our method's performance remains highly stable across the tested range of $K$ values.
\begin{figure*}[t]
  \includegraphics[width=\textwidth]{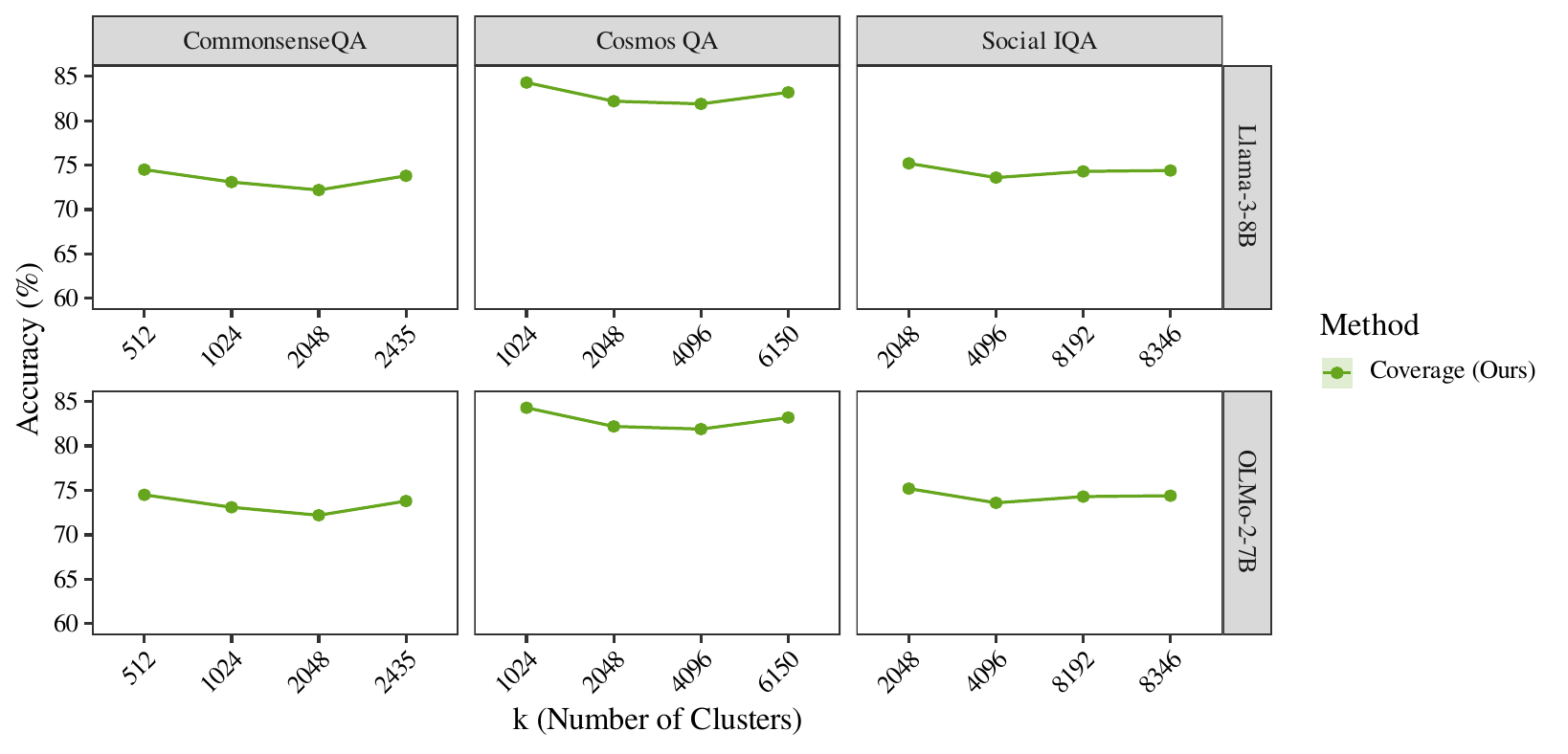}
  \centering
  \caption{Per-dataset and average accuracy across
different values of the clustering parameter $k$ at 25\% selection. }
  \label{fig:accuracyWithDifferentK}
\end{figure*}
\subsection{Implementation Details}\label{sec:hyperparameters}
For large-scale model fine-tuning, we use Huggingface Accelerate~\citep{accelerate} and utilize Zero3 optimizer in DeepSpeed~\citep{DBLP:conf/kdd/RasleyRRH20}. We find  the optimal values through hyperparameter search over learning rates \{1e-05,2e-05\}, batch size \{64, 128\} and, number of epochs  \{2,3\}. Based on the performance, we set learning rate as 2e-05 and batch size as 128 and number of epochs as 3. Models are
trained on selected subsets using the same hyperparameters that are used for training 100\% of the data. This follows from previous works like ~\citep{maharana2023d2pruningmessagepassing, anand-etal-2023-influence, sorscher2022beyond}

\subsection{Effect of layer on clustering}\label{sec:layer_experiment}
In the main paper, we used the embeddings from the last layer for k-means in clustering in \Cref{sec:coverage_based_method}. To investigate if intermediate layer embeddings can be more helpful, we repeat our experiments using embeddings from the middle layer while selecting 10\% of the data. ~\Cref{fig:layerAblation} shows the performance when using the last layer and the middle layer. Across models and evaluation benchmarks, intermediate-layer embeddings did not provide an improvement: final-layer embeddings consistently achieved higher average accuracy.
\begin{figure*}[t]
  \includegraphics[width=\textwidth]{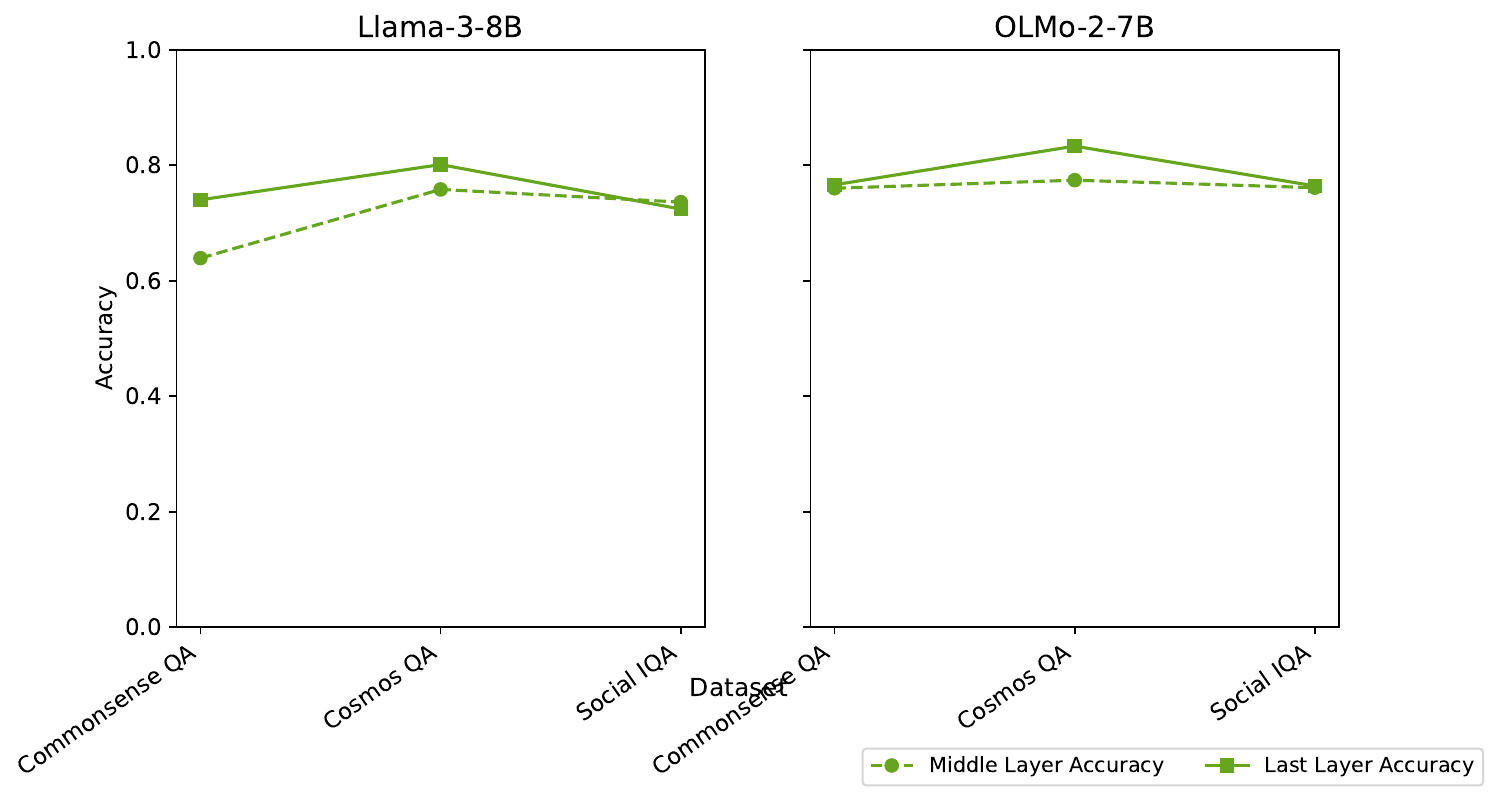}
  \centering
  \caption{Comparative performance of 10\% selected data on the \llama{} and \olmo{} models across all the datasets. Accuracy on CommonsenseQA, CosmosQA, and Social IQA  using Last Layer vs. Middle Layer embeddings in the clustering step of our coverage based method. }
  \label{fig:layerAblation}
\end{figure*}

\end{document}